\pdfoutput=1

\newcommand{\frameworkname}{\textsc{CritiCS}}
\newcommand{\leaderGPT}{leader}
\newcommand{\evalGPT}{evaluator}
\newcommand{\RePlan}{\textsc{CrPlan}}
\newcommand{\ReSent}{\textsc{CrText}}
\documentclass[11pt]{article}

\usepackage{acl}

\usepackage{times}
\usepackage{latexsym}
\usepackage{graphicx} 
\usepackage[T1]{fontenc}
\usepackage{amsmath,amssymb,amsfonts}
\usepackage{amsthm}
\usepackage{mdframed}
\usepackage{array}
\usepackage{algorithm}
\usepackage{algpseudocode}
\usepackage{booktabs}
\usepackage{hyperref}
\usepackage{multirow} 
\usepackage{tabularx}
\usepackage{xcolor}
\usepackage{soul}
\usepackage{enumitem}

\definecolor{plan}{RGB}{211, 250, 197}
\definecolor{textCRIT}{RGB}{250, 249, 178}
\definecolor{Negative}{RGB}{250, 207, 197}
\newcommand{\plan}[1]{ 
  \begingroup
  \unskip
  \sethlcolor{plan}
  \textbf{\hl{#1}}
  \unskip
  \endgroup
}
\newcommand{\textCRIT}[1]{
  \begingroup
  \unskip
  \sethlcolor{textCRIT}
  \textbf{\hl{#1}}
  \unskip
  \endgroup
}
\newcommand{\negative}[1]{
  \begingroup
  \unskip
  \sethlcolor{Negative}
  \textbf{\hl{#1}}
  \unskip
  \endgroup
}

\usepackage[utf8]{inputenc}

\usepackage{microtype}

\usepackage{inconsolata}

\title{Collective Critics for Creative Story Generation}

\author{Minwook Bae\quad Hyounghun Kim \\
  Artificial Intelligence Graduate School, UNIST \\
  \texttt{\{minwook09, h.kim\}@unist.ac.kr} \\}

\begin{document}
\maketitle

\begin{abstract}
Generating a long story of several thousand words with narrative coherence using Large Language Models (LLMs) has been a challenging task. Previous research has addressed this challenge by proposing different frameworks that create a story plan and generate a long story based on that plan. However, these frameworks have been mainly focusing on maintaining narrative coherence in stories, often overlooking creativity in story planning and the expressiveness of the stories generated from those plans, which are desirable properties to captivate readers' interest.
In this paper, we propose Collective \textbf{Criti}cs for \textbf{C}reative \textbf{S}tory Generation framework (\frameworkname{}), which is composed of plan refining stage (\RePlan) and story generation stage (\ReSent), to integrate a collective revision mechanism that promotes those properties into long-form story generation process. Specifically, in each stage, a group of LLM critics and one leader collaborate to incrementally refine drafts of plan and story throughout multiple rounds. 
Extensive human evaluation shows that the \frameworkname{} can significantly enhance story creativity and reader engagement, while also maintaining narrative coherence. Furthermore, the design of the framework allows active participation from human writers in any role within the critique process, enabling interactive human-machine collaboration in story writing.\footnote{Our code are publicly available at \url{https://github.com/EMNLP-2024-CritiCS/Collective-Critics-for-Creative-Story-Generation}}
\end{abstract}

\section{Introduction} 
\begin{figure*}[t]
    \centering
    \includegraphics[width=2.0\columnwidth]{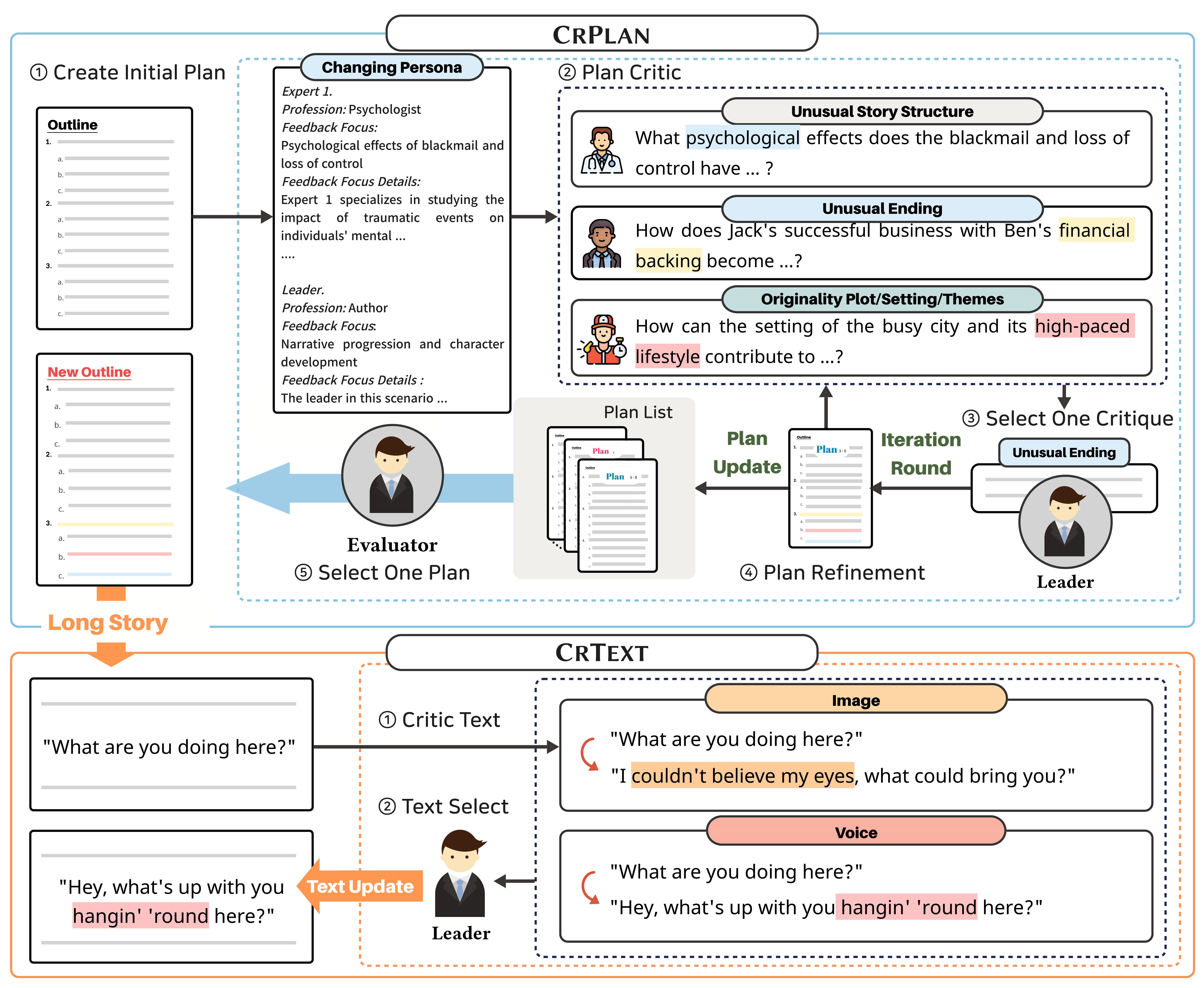}
    \caption{The framework comprises two stages: \RePlan{} and \ReSent{}. \RePlan{} involves five phases: creating a story plan from a premise, reviewing the plan with critics' persona-driven perspectives, selecting a critique for revision by a leader, storing the revised plan, and choosing a plan for further development by an evaluator. Personas of critics are created based on the themes or content of the narratives, which helps in generating detailed and contextually relevant critiques. Please refer to Appendix~\ref{abl:persona_adpative} critiques and refines long story text based on creative criteria, with a leader selecting and improving the best expressions.}
    \label{fig:model}
    \vspace{-5pt}
\end{figure*} 
Recent progress in Large Language Models (LLMs) has enabled the automatic generation of long-form stories containing several thousand words. This advancement has led to the proposal of various frameworks~\cite{yang-etal-2022-re3,yang-etal-2023-doc, zhou2023recurrentgpt} that are capable of generating stories maintaining narrative coherence. In pursuit of this capability, many frameworks adopt a two-step approach~\cite{fan-etal-2018-hierarchical, yao2019plan, fang2021outline, tian-peng-2022-zero}: (a) creating a story plan; and (b) generating a detailed story based on the plan. To explore further, some works have made effort to improve structure of story plan~\cite{wang-etal-2023-improving-pacing} and refine the writing style~\cite{kong-etal-2021-stylized,zhu-etal-2023-storytrans}.
Despite their ability to produce consistent and coherent text, these works frequently fall short in crafting creative long-range narratives that captivate human interest, such as stories with emotional flow, flashback structures, and other engaging elements~\cite{zhang-etal-2021-trading, huang-etal-2023-affective}. 

Therefore, we propose Collective \textbf{Criti}cs for \textbf{C}reative \textbf{S}tory Generation (\frameworkname{}), a novel framework for long-form story generation that integrates collaborative critique into the plan-to-story framework. Our framework focuses on creatively refining story plans (\RePlan{}) and enhancing the expressiveness of stories (\ReSent{}). Inspired by studies demonstrating that collaborative work significantly enhances creativity through the sharing and integration of knowledge and ideas from various fields~\cite{mamykina2002collaborative, barrett2021creative}, \frameworkname{} adopts a collaborative critique approach.
Specifically, in each stage, multiple LLM critics evaluate the draft and provide suggestions for improvement based on criteria (e.g., uniqueness of the narrative flow and vividness of the generated text) which is designed for assessing creativity~\cite{mozaffari2013analytical}. Then, a \leaderGPT{} selects a critique that best helps refine the draft. This revision process iterates through multiple rounds to produce a final plan and story~(Figure~\ref{fig:model}).  

In \RePlan{}, the critics focus on infusing originality into the story plan (e.g., unexpected shifts in the story's atmosphere, twisted endings). To do so, critics are endowed with personas that are adaptive to a given storyline, which helps make clear and diverse suggestions to enhance the story's creativity and coherence.
\ReSent{} is designed to improve the expressiveness of stories. It refines the narrative to incorporate unique expressions and vivid descriptions. This includes comic book onomatopoeias (e.g., `\textit{swooosh}'), figurative phrases (e.g., `\textit{silent twilight}'), and detailed sensory experiences (e.g., a `\textit{buzzing market}' or the `\textit{sharp tang of sea air}').

Extensive human evaluation demonstrates that our \frameworkname{} produces more creative and thus interesting plan, and generates expressive story. To be specific,
\frameworkname{} outperforms the state-of-the-art method on the metrics of creativity and interestingness by a large margin. 
Additionally, detailed analyses indicate that our design choices for \frameworkname{} (e.g., applying multiple criteria, role of leader critics, and bestowing personas) 

In addition, \frameworkname{} facilitates interactive human-machine collaboration in writing by enabling humans to participate as any of the players within the framework, refining stories according to their preferences.

The main contributions of this study are:
\begin{enumerate}
    \item We propose a framework named \frameworkname{}, which is designed to enhance story plans and text expressiveness creatively through collective critiques based on criteria of creativity.
    
    \item Extensive human evaluation and ablation studies demonstrate that \frameworkname{}  effectively creates stories with creativity and coherence, employing key components such as diverse criteria, leaders, and personas.

    \item \frameworkname{} supports interactive writing, enabling human users to intervene in critiques and actively refine stories.
    
\end{enumerate}

\section{Related Work}
\paragraph{Long-form Story Generation.}
In prior research, the focus of automatic story generation is to create short stories composed of several completed sentences~\cite{roemmele-gordon-2018-linguistic, Ammanabrolu2020StoryRealization,wang-etal-2022-co}. With the development of LLMs, studies on story generation have shifted their focus toward generating longer narratives of several thousand words, maintaining narrative coherence~\cite{yang-etal-2022-re3,yang-etal-2023-doc,zhou2023recurrentgpt}. Those works have tried to sustain coherence in longer stories by employing story planning techniques such as hierarchical generation~\cite{fan-etal-2018-hierarchical, fan-etal-2019-strategies}. This approach involves first crafting a broad story outline at a high level, which serves as the foundation for the detailed narrative development. Such hierarchical story plans typically consist of outlines with around 1K tokens, allowing for the creation of longer stories without losing coherence.
Recent research focuses on creating robust story plans (e.g., story plans with high narrative coherence, story plans with appropriately placed event scenes)~\cite{Ammanabrolu2020StoryRealization,zhu2023endtoend,wang-etal-2023-improving-pacing} and transferring plain text into literary expressions~\cite{kong-etal-2021-stylized,zhu-etal-2023-storytrans,huang-etal-2023-affective}. However, those works fall short in devising unusual, original stories that engage the reader's interest. To address these issues, in this study, we focus on the creativity of stories.
\paragraph{Creative Writing.}
The assessment of creativity in creative writing has been persistently discussed among researchers. Traditional approaches to assessing creativity have relied on criteria like novelty and unexpectedness~\cite{barron1955disposition,rhodes1961analysis,blomer2011assessment}. However, these approaches are abstract, as they typically rely on broad standards for evaluation. The need for specific criteria to assess the writing has led to rubric-based methods using the traits of creative writing~\cite{arter2001scoring, young2009imagine,mozaffari2013analytical}, generally assessing creativity in three aspects~\cite{mozaffari2013analytical}: \textit{Image}, \textit{Voice}, and \textit{Originality}. In this study, we adopt those criteria to give more detailed guideline for critcs to refine stories.
\paragraph{Collaborative Writing.}
Research on collaborative writing as a method to produce creative authoring is actively ongoing~\cite{csikszentmihalyi1997flow, mamykina2002collaborative,paulus2012collaborative}. Compared to individual writing, collaborative writing can lead to more divergent works by sharing different knowledge~\cite{bayle1998putting,resnick2005design,sie2009knowledge,bitter2011new}. The more diverse the backgrounds, skills, and experience levels of the collaborators, the more varied the knowledge shared, enhancing the creativity of the writing. Inspired by this, we design~\frameworkname{} to accommodate multiple critics, each with unique characteristics, to increase diversity.

\paragraph{Interactive Agents.}
Recently, there has been a strong focus on developing interactive agents powered by LLMs~\cite{OpenAI2023,team2023gemini, anil2023palm, openai2024gpt4}. These agents aim to enhance collaboration, enabling them to share information and solve complex problems together. Earlier studies~\cite{DBLP:conf/iclr/YaoZYDSN023,du2023improving,gou2024tora,sun2024corex,chen2024reconcile} focused on improving the reasoning capabilities of LLM agents through interactions. However, recent findings~\cite{DBLP:journals/corr/abs-2310-01798,wang2024rethinking} suggest that their reasoning abilities cannot improve solely through interaction without external feedback.
On the other hand, current research has shifted towards tasks with various possible solutions, such as code generation~\cite{chen2023teaching,hong2023metagpt,DBLP:journals/corr/abs-2310-02003} and embodied AIs~\cite{DBLP:conf/iccv/SongSWCW023,wang2023voyager,shek2024lancar,chen2024pcabench}, where agents cooperate to achieve specific goals. 
Our research utilizes interactive agents to build divergent critique processes for creative long-story generation. Integrating human and agent participants enriches narratives with diverse perspectives, making long stories more engaging and creative.
\section{Collective Critics for Creative Story Generation}
\frameworkname{} is composed of two stages (\RePlan{} and \ReSent{}) that utilize the capabilities of LLM to write creative stories. Our framework enhances creativity at both the micro level (sentence expressiveness) and the macro level (story structure, themes, endings, etc.) To achieve this, we incorporate the critique process based on creative criteria~\cite{mozaffari2013analytical} into our framework. Although we commonly use criteria for assessing the creativity of stories in literature, our framework is designed to accommodate a variety of other criteria as well. 
\subsection{\RePlan{}}
In \RePlan{}, three critics assess the creativity of story plans based on each of three criteria: `\textit{Original theme and background setting}’, `\textit{Unusual story structure}’, and `\textit{Unusual ending}’. This assessment ensures that it imbues diverse elements of the story with a sense of novelty. Table~\ref{exam:CRPLAN} shows an example of refined plan wherein the protagonist's loneliness is anthropomorphized, enhancing the uniqueness of the narrative setting.

First, each of the three critics offers a suggestion to enhance the draft plan, using their unique expertise. Next, the leader critic evaluates the three suggestions, ranks them, and selects the one it believes to be the best. The chosen suggestion is then applied to refine the plan.  
For detailed critiques, the three critics are given personas with expertise relevant to the story plan. Meanwhile, the leader is embodied as a literature editor or creative writing expert, tasked with reconciling any conflicts.
This whole process is repeated over several rounds, after which a plan evaluator reviews the candidate plans from each round to select the final one that effectively balance creativity with coherence. Selection criteria, evaluation, hyper-parameters, and prompts are detailed in Appendix~\ref{prompt:all}.
\begin{algorithm}[t]
\small
\caption{\small \frameworkname{}: \RePlan{} Stage}
\label{alg:Retext}
\textbf{Input:} iteration $I$, story\_plan $T$, critics $L$ \\
\textbf{Output:} \emph{best\_plan}
\begin{algorithmic}[1]
    \State {story\_list $\gets$ [ ]}
	\For {$i=0,1,\ldots,I$}
		\State $c_{\text{O}}, c_{\text{S}}, c_{\text{E}} \gets L(T_i)$	\Comment{$c_{n}$ : criteria and persona}
        \State $C_s \gets Leader(c_{\text{o}}, c_{\text{s}}, c_{\text{e}})$
        \State $T_i \gets \textsc{Refine}(T_p, C_s)$
        \State story\_list.append$(T_i)$
	\EndFor \\
    \Return $Evaluator$(story\_list)
\end{algorithmic} 
\end{algorithm}
\begin{algorithm}[t]
\small
\caption{\small\frameworkname{}: \ReSent{} Stage}
\label{alg:Replan}
\textbf{Input:} long\_text $T$, critics $L$ \\
\textbf{Output:} \emph{Refined\_Text}
\begin{algorithmic}[1]
    \State $c_{\text{I}}, c_{\text{V}} \gets L(T_i)$ 
    \State $C_s \gets Leader( c_{\text{I}}, c_{\text{V}})$ \Comment{$c_n$: criteria}
    \State $T_i \gets \textsc{Refine}(T_p, C_s)$
    \State $T_p \gets T_i$ \\
    \Return $T_p$ 
\end{algorithmic}
\end{algorithm}
\subsection{\ReSent{}}
In \ReSent{}, two critics review a story created based on the plan from \RePlan{}, using  each of two criteria focused on the expressiveness of creative text: \textit{Image} and \textit{Voice}. As an outcome of this process, Table~\ref{exam:Text} shows an example of refined text in which the verb `raised' is replaced with `arched,' making the phrase more distinctive.

Image indicates the degree to which a reader is provoked with vivid mental imagery. This includes visual images, sounds, smells, bodily sensations, and emotions (e.g, soft glow emitted from the moon, eerie shadows). Voice means the extent to which an author has succeeded in creating a unique and recognizable writing style (e.g., written in a horror story style, slang words, informal language like `\textit{lol}'). Similar to \RePlan{}, the \leaderGPT{} choose one of two suggestions for refining expressions.

Unlike in \RePlan{}, the critics have no persona because Image and Voice criteria provide clear instructions for expression modification, eliminating the need for personas to anchor the story theme. Also, there is no evaluator because different sentences are revised in each round. Please see Figure~\ref{fig:model} (bottom) for the whole process. The criteria for selection and evaluation and prompts used in this stage are detailed in Appendix~\ref{prompt:all}.
\begin{table}[t]
\small
\begin{tabular}{m{0.95\linewidth}}
\toprule
\multicolumn{1}{l}{\textbf{Initial Plan}} \\
\midrule
\midrule
1. Aimee Kincaid goes home after a long day at work only to find her apartment empty and her loneliness amplified by the silence. \\
\textCRIT{Characters: Aimee Kincaid, Kyle Johnson} \\
\\
\quad a. Aimee Kincaid comes home to her empty apartment after a long day at work ... \\
\midrule
\multicolumn{1}{l}{\textbf{Critiques}}\\
\midrule
\midrule
Critique: What if Aimee's loneliness is personified as a physical entity that she can interact with?\\
\\
Why: This critique introduces a unique twist by personifying loneliness and turning it into a tangible character. It adds a fantasy element to the story and deviates from the traditional romance storyline, making it quite different, novel, and innovative. \\ 
\midrule
\multicolumn{1}{l}{\textbf{Refined Plan (Personalized)} } \\
\midrule
\midrule
1. Aimee Kincaid goes home after a long day at work only to find her apartment empty and her loneliness manifests as a physical entity ...  \textCRIT{Characters: Aimee Kincaid,} \plan{Loneliness(personified)}\\
...\\
\quad    b. \plan{Loneliness interacts with Aimee}, amplifying her feelings of isolation and despair. Scene:  Characters: Aimee Kincaid, Loneliness(personified) ...\\
\bottomrule
\end{tabular}
\vspace{-7pt}
\caption{Example of enhancing story setting originality by personifying the characters' loneliness through \RePlan{}. Full story plan in Appendix~\ref{example:all}.}
\label{exam:CRPLAN}
\vspace{-7pt}
\end{table}
\begin{table}[t]
\small
\begin{tabular}{m{0.95\linewidth}}
\toprule
\multicolumn{1}{l}{\textbf{Initial Text}} \\
\midrule
\midrule
... He leaned against the moss-covered trunk of a nearby tree, peering at Jonathan with curious eyes. \textCRIT{Jonathan raised an eyebrow}. "What do you mean?" ...\\
\midrule
\multicolumn{1}{l}{\textbf{Refined Text (Unique Verb : "arched")}} \\
\midrule
\midrule
...He leaned against the moss-covered trunk of a nearby tree, peering at Jonathan with curious eyes. \plan{Jonathan arched an incredulous eyebrow}. "What do you mean?" ...\\
\bottomrule
\end{tabular}
\caption{Example of enhancing text expressiveness (Voice) by using ``arched'' instead of ``raised'' and ``incredulous'' for depth, emphasizing Jonathan's surprise.}
\label{exam:Text}
\vspace{-10pt}
\end{table}

\section{Experiments}
\label{exp:main}
\frameworkname{} focuses on enhancing the story's creativity while preserving narrative coherence. To assess the effectiveness of this approach, we conduct experiments that evaluate two stages of our framework, \RePlan{} and \ReSent{}. Specifically, we qualitatively compare 300 plans and stories generated by \frameworkname{} with those from the state-of-the-art long-form story generation framework, the DOC pipeline~\cite{yang-etal-2023-doc}. For fair comparisons, we employ the ChatGPT (gpt-3.5-turbo) to generate plans and stories for both DOC and our \frameworkname{}.
To ensure high quality and fairness in our experiments, we adopt a human evaluation, where three inter-annotators assess each pair of samples from both methods and determine which is superior. Additionally, to demonstrate the versatility of our framework, we conduct streamlined human evaluations using GPT-4~\cite{openai2024gpt4} as an LLM backbone and Re3~\cite{yang-etal-2022-re3} as a baseline, confirming its effective operation across a variety of LLM backbones and baselines. Please refer to Appendix~\ref{exp:other_baseline} for these results.
\subsection{\RePlan{} Evaluation}
\paragraph{Generation Setting.}
To generate initial story plans, we use 300 premises from the DOC~\cite{yang-etal-2023-doc}. We regulate the number of rounds to 3. Experimental findings suggest that more revision rounds leads to significant changes in the story, which can be interpreted as heightened inventiveness. However, this excessive inventiveness significantly detracts from the narrative's coherence. For more details, please refer to Section \ref{abl:Iter}.
\vspace{1pt}
\par
\noindent\textbf{Metrics.}
We conduct pairwise comparison for human evaluation following~\citet{yang-etal-2023-doc}. Additionally, we employ metrics to measure the creativity of the stories from~\citet{zedelius2019beyond}. Definitions are as follows:
\begin{itemize}
\itemindent=-10pt
    \item Interesting: This evaluates how effectively the story plan engages and captivates readers.
    \item Coherence: This evaluates how logically organized and interconnected the story plans are.
    \item Relevant: This evaluates how well the narrative themes of the story plan adhere to the premise.
    \item Creative: This evaluates the story plan's originality and inventiveness, focusing on its fresh perspective compared to typical narratives.
\end{itemize}
Annotators are asked to choose a good one based on each of the metrics. If they think both are good or bad, they are allowed to select `Both are about equally good' or `Neither is good'. Details of the questionnaire can be found in Appendix~\ref{exp:eval_detail}.
\vspace{3pt}
\par
\noindent\textbf{Result.}
As shown in Table~\ref{tbl:Hueval}, \RePlan{} shows superior performance on three metrics by a large margin, scoring slightly higher on Relevance. This signifies the effectiveness of \RePlan{} in enhancing the creativity of stories while maintaining coherence. Additionally, we observe that as stories are revised by critiques, the coherence of the story also improves. This improvement is attributable to the detailed critiques generated through the persona-driven critique, which facilitates sensitive modifications to the existing narrative content. Such detailed storytelling ensures that the robust narrative without omissions, unfolding clearly and conveying strong coherence to readers. Please refer to Section~\ref{abl:Per} for an analysis of how the persona-driven critique improves narrative coherence.
\begin{table}[t]
\renewcommand{\arraystretch}{1.3}
\centering
\resizebox{0.999\linewidth}{!}{
    \begin{tabular}{@{}lcccc@{}}
    \toprule[1.3pt]
      & \multicolumn{4}{c}{\textbf{Human Evaluation}} \\
    \cline{2-5}
    \textbf{Model} & \textbf{Interesting\textsuperscript{\textuparrow}} & \textbf{Coherent\textsuperscript{\textuparrow}} & \textbf{Creative\textsuperscript{\textuparrow}} & \textbf{Relevant\textsuperscript{\textuparrow}} \\  
    \midrule
    DOC & 57.56 & 67.33 & 57.33 & 95.11 \\
    \RePlan{} & \textbf{85.00} & \textbf{77.89} & \textbf{84.33} & 96.00 \\ [0.5ex]
    Fleiss' Kappa & 0.231 & 0.399 & 0.518 & 0.192\\
    \bottomrule[1.5pt]
    \end{tabular}
}
\caption{Results of human evaluation for 300 pairwise story plan comparisons of \RePlan{} vs  baseline. Bold indicates $p < 0.05$ significance. Most inter-agreements are fair to moderate. Kappa score for `Relevant' is low due to data bias, as most annotators rated both good.}
\label{tbl:Hueval}
\vspace{-12pt}
\end{table}
\subsection{\ReSent{} Evaluation}
\paragraph{Generation Setting.}
We initially generate 300 long stories following the process from~\citet{yang-etal-2023-doc}, using story plans from \RePlan{}. In each round, we randomly select one sentence from each long story and refine it through \ReSent{}, then pair-wisely compare with the stories generated by DOC pipeline. We set the number of rounds to 3.
\vspace{3pt}
\par
 \noindent\textbf{Metrics.}
We utilize the evaluation criteria as used in~\citet{kong-etal-2021-stylized} and~\citet{zedelius2019beyond}. These criteria assess whether each refined sentence appropriately fits within the narrative in terms of coherence, maintaining consistent writing style, and the sentence relevance in the narrative context. In addition to these two criteria, we also include measures of the sentence's creativity and interestingness it evokes in readers as part of our evaluation criteria. The detailed criteria are as follows:
\begin{itemize}
\itemindent=-10pt
    \item Coherence: This evaluates sentence relevance and their causal and temporal dependencies, ensuring a logical and smooth narrative flow.
    \item Writing Style Consistency: This evaluates whether the style of the given sentence is consistent with the overall context of the story.
    \item Interesting: This evaluates whether the expression in the sentences is rich and engaging, contributing interesting elements to the story.
    \item Creative: This evaluates whether the sentences include unique or rich expressions that differ from those in a typical story.
\end{itemize}
\par
\noindent\textbf{Result.}
As shown in Table~\ref{tbl:TextHE}, \ReSent{} demonstrates significantly higher performance on two metrics, interestingness and creativity, also showing slightly better performance for the others. This verifies the effectiveness of \ReSent{} in enhancing the expressiveness of text creatively and engagingly while maintaining the writing style and coherence.
\begin{table}[t]
\renewcommand{\arraystretch}{1.3}
\centering
\resizebox{0.999\linewidth}{!}{
    \begin{tabular}{@{}lcccc@{}}
    \toprule[1.3pt]
      & \multicolumn{4}{c}{\textbf{Human Evaluation}} \\
    \cline{2-5}
    \textbf{Model} & \textbf{Interesting\textsuperscript{\textuparrow}} & \textbf{Coherence\textsuperscript{\textuparrow}} & \textbf{Consistency\textsuperscript{\textuparrow}} & \textbf{Creative\textsuperscript{\textuparrow}} \\  
    \midrule
    DOC & 69.89 & 69.11 & 76.89 & 70.67 \\
    \ReSent{} & \textbf{80.00} & 71.89 & 80.00 & \textbf{89.33} \\ [0.5ex]
    Fleiss' Kappa & 0.382 & 0.338 & 0.342 & 0.320\\
    \bottomrule[1.5pt]
    \end{tabular}
}
\caption{Results of human evaluation for 300 pairwise story expressiveness comparisons of \ReSent{} against the baseline. Bold indicates statistical significance with $p < 0.05$. Most inter-agreements are fair.}
\label{tbl:TextHE}
\vspace{-15pt}
\end{table}
\vspace{-10pt}
\section{Analysis}
We explore the effectiveness of our design choices for \frameworkname{} through an ablation study of its settings. To ensure a comprehensive evaluation, we utilize various methods including streamlined human evaluation, automatic assessments with GPT-4—widely adopted in research for evaluating story plans~\cite{wang-etal-2023-improving-pacing,zhu2023endtoend,you2023eipetext}—and diverse examples. To verify the reliability of GPT-4's automatic evaluation and its alignment with human judgments, we report the inter-agreement between the human evaluation results from Section~\ref{exp:main} and the automatic evaluation results. In Section~\ref{exp:main}, three annotators evaluate 300 stories, and we calculate the Cohen's Kappa score between each annotator's evaluations and GPT-4's automatic evaluation of the same stories. The results show a fair to moderate level of agreement between the human evaluators and GPT-4. Please refer to Appendix~\ref{exp:autoamtic_anno} for details on the reliability of GPT-4’s automatic evaluation.
\subsection{Analysis of Criteria.}
\par
\noindent\textbf{Diversity of Criteria.}
\label{abl:mod}
\begin{table}[t]
\small
\begin{tabular}{m{0.95\linewidth}}
\toprule
\multicolumn{1}{l}{\textbf{Initial Story Plan}}\\
\midrule
\midrule
1. John \textCRIT{discovers the package of cash} and visits Abe every day to share his lunch, and they develop a bond. \\...\\
2. With Abe's guidance, John starts an art therapy group\\ ...\\
3. While John's mental health improves, he becomes closer to Alyssa and Jake, who support him in exploring his artistic side and his aspirations. ...\\
\midrule
\multicolumn{1}{l}{\textbf{Refined Plan: `Original Themes’ Criterion}}\\
\midrule
\midrule
1. John discovers the package of cash and wrestles with the \negative{ethical dilemma}, even as he grows closer to Abe.
\\...\\
2. Learning about the~\negative{digital world}, John integrates technology with art therapy and starts a group in his school, forming meaningful connections\\...\\
3. John's renewed connection with Alyssa and Jake strengthens amidst the success of his \negative{digital-art therapy group.} ...\\
\midrule
\multicolumn{1}{l}{\textbf{Refined Plan: Three Criteria (Flashback Structure)}}\\
\midrule
\midrule
1. John is now leading a successful \plan{art therapy group} at the school \\...\\
3. \plan{Flashbacks reveal John's past}, depicting his \plan{discovery of a package of cash} one day while walking his dog Max in the forest behind his house. ...\\
\bottomrule
\end{tabular}
\caption{Example of story plan comparison in \RePlan{}: single-criterion vs. three-criteria critiques. The three-criteria critique addresses thematic content and shifts event structure with flashbacks, while the single-criterion critique focuses on a single theme, like ethical dilemmas or the digital world.}
\label{anlys:onecritic}
\vspace{-6pt}
\end{table}
To assess the effect of criteria diversity, we conduct a comparative study. In one setting, we assign the same criterion to all critics, while in the other, each critic is given different criteria. Table~\ref{anlys:onecritic} shows examples comparing the refined story plans through single criterion and three criteria critiques in \RePlan{}. The critique based on a single criterion reveals limitations in creating various story flows, such as changes in narrative structure, changing only narrative settings, such as ethical dilemmas or digital worlds.

Conversely, revisions made through a three-criteria critique contribute to various aspects of storytelling originality, including unique story materials and narrative structures. The initial plan might feature a straightforward chronological narrative where John walks down a street, finds cash, and returns it to Abe. However, the refined plan evolves into a more complex and engaging narrative: it starts with John leading a therapy group as an adult. It then cleverly transitions through flashbacks to recount his past with Abe, showing a unique narrative structure and themes. This approach enriches the story's depth and significantly demonstrates the potential of three-criteria critiques to enhance narrative creativity and complexity.

Single-criterion critiques in \ReSent{} might not adequately meet the needs for appropriate improvement. For example, critiquing a dialogic sentence like ``\textit{But... why me?}'' using only the Image criterion tends to yield merely a detailed description without enhancing its expressiveness such as ``\textit{But... why have I been chosen for this?}'' However, incorporating the Voice criterion can transform the sentence into something unique like ``\textit{Umm, but... why me, exactly?}'' by adding filler word, thus enhancing the text's expressiveness.
The full plan and story are in Appendix \ref{ablappend:one critic}.
\vspace{3pt}
\par
\noindent\textbf{Different Criteria.}
To demonstrate our framework's adaptability across various criteria, we develop specific metrics for creative characters (Dynamic Development) and sentence structures (Inverted and Non-linear Structure), applying these to each \RePlan{} and \ReSent{} stage, respectively. This shows the framework's ability to adapt to diverse criteria, enabling users to steer the story's creative direction according to their preferred standards. Please refer to Appendix~\ref{abl:vairious_critieria} for details on various criteria and their application.
\subsection{Roles of Leader.}
\label{abl:evalmodules}
\begin{table}[t]
    \small
    \begin{tabular}{m{0.95\linewidth}}
    \toprule
    \multicolumn{1}{l}{\textbf{Without Leader Critique}} \\
    \midrule
    \midrule
    ...\\
    \quad    c. Despite the treachery, the \negative{remaining group pushes forward without Jake}, their resolve doubling.\\
    3. As \negative{Jake and his friends} delve deeper into the conspiracy, the diamond heist is revealed to be emblematic\\...\\
\quad    a. Their investigations expose a vast conspiracy, ...\\
    \toprule
    \multicolumn{1}{l}{\textbf{With Leader Critique (Complex Character Relation)}}\\
    \midrule
    \midrule
    ...\\
	a. Jake, the group's leader, has deep feelings for Sarah, but \plan{Sarah is romantically caught between Jake and Tom}, causing tension amongst them.\\
	...\\

3. \plan{Jake gets captured} and held hostage at the city museum, causing panic within the team.
\\
\quad a. In Jake's absence, Tom shines as interim team leader, further escalating tensions between him and Jake. \\...\\
4. ...\\	
\quad    b. Sarah and Tom, fueled by their need to \plan{save Jake} and to uphold what's right for their city, put their feelings aside, setting the stage for an intense final showdown. ...\\
    \bottomrule
    \end{tabular}
    \caption{Example of story plan comparison between without-leader-critique and with-leader critiques in \RePlan{}. With-leader critiques ensure the maintenance of narrative coherence amidst complex character relations.}
    \label{anlys:leader}
    \vspace{-10pt}
\end{table}
We generate plans and stories through our framework with and without leaders to investigate the implications of their absence.

Table~\ref{anlys:leader} provides comparative examples of a story plan refined with a leader involved and the other generated without a leader,  while all three critics are engaged in \RePlan{}. In the leaderless critique example, the revision portrays both Jake's treachery and his abrupt collaboration with the group on investigating political secrets in the subsequent scene, without offering any explanation, thereby revealing a narrative contradiction. 
In contrast, the revision from the critique with a leader demonstrates that narrative consistency is maintained despite the complexity of character relationships. Initially, the story presents a triangular relationship between Sarah, Jake, and Tom. Following Jake's abduction, the narrative continues to preserve Sarah's enduring love for Jake in the latter part of the story.

In the absence of leader in \ReSent{}, critiques may not always align with the text's intended context. For instance, consider a dialogic sentence like ``\textit{I never thought anyone would understand.}'' Without a leader's selection, choosing an Image critique might lead to overly detailed revisions, such as ``\textit{I always believed my thoughts were whispers, too faint for anyone to truly hear.}'' which may not suit the dialogic nature of the original sentence. In contrast, when the leader selects Voice critique, the sentence can be aptly refined to ``\textit{Never did I reckon anyone would get it, y'know.}'' enabling a suitable modification. The full plan and story can be found in Appendix~\ref{ablappend:leader}.
\vspace{-4pt}
\subsection{Roles of Persona.}
\begin{figure}[t]
    \centering
    \includegraphics[width=0.99\columnwidth]{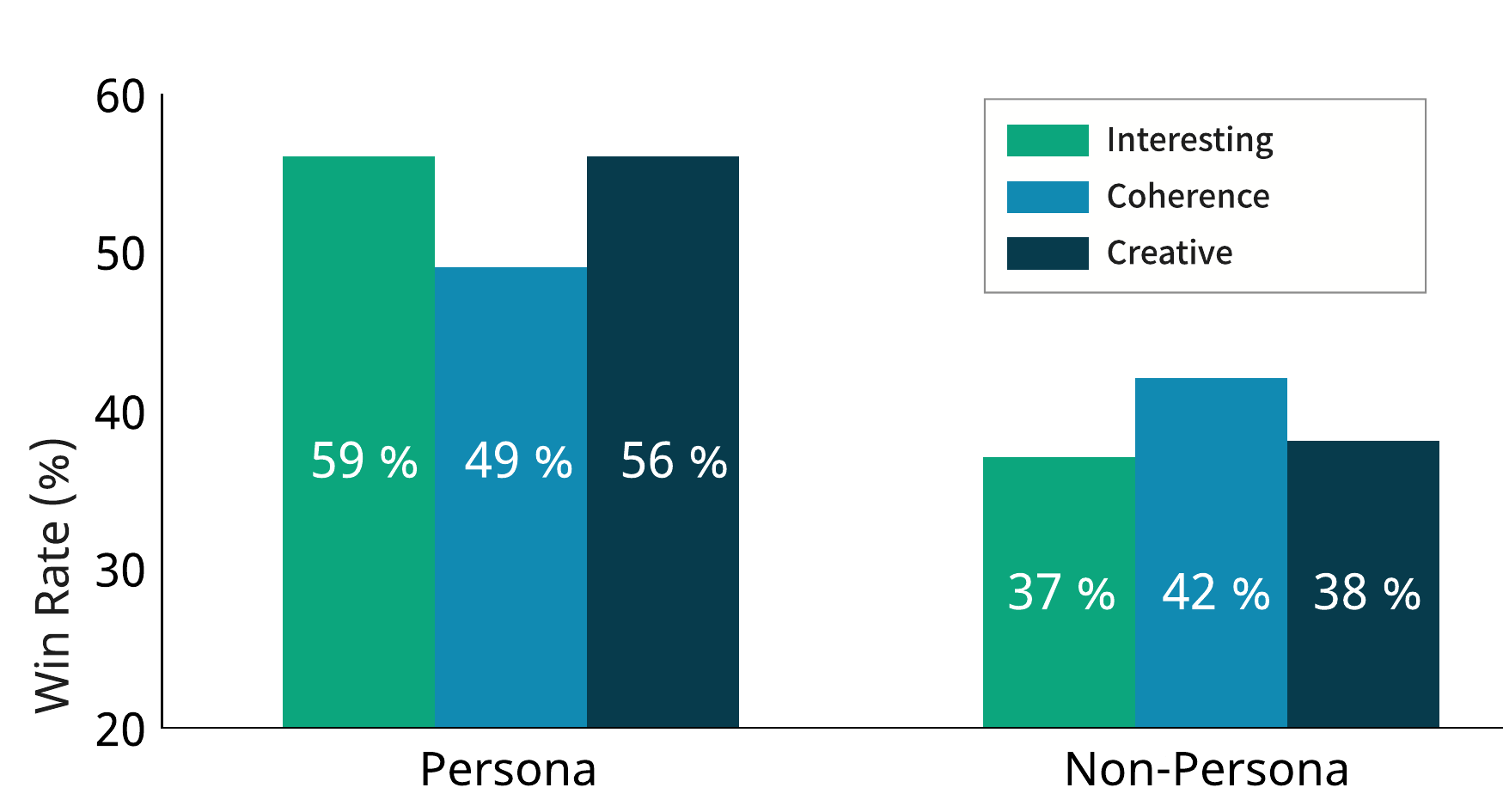}
    \caption{Pairwise story plan comparisons using GPT-4: Non-Persona-Critics vs. Persona-Critics.}
    \label{fig:persona_anlaysis}
    \vspace{-10pt}
\end{figure}
\label{abl:Per}  
Table~\ref{anlys:persona} shows the distinctions between suggestions from critics with and without personas in \RePlan{}. The examples reveal that critics without personas often provide general story improvement suggestions, such as enhancing the narrative structure or reorganizing the storyline. They may also deliver critiques on themes unrelated to the central narrative, for instance, commenting on alternate dimensions or parallel universes for a story focused on new energy development. These broad and irrelevant critiques undermine narrative coherence and hinder the creation of a diverse range of critiques. The full critiques can be found in Appendix~\ref{ablappend:non-persona}.

We also conduct an automatic evaluation of 100 story plan pairs using GPT-4, based on evaluation metrics in Section~\ref{exp:main}. For details on the automatic evaluation methodology, please refer to Appendix~\ref{exp:eval_detail}. The results in Fig~\ref{fig:persona_anlaysis} show that personas foster diverse critical perspectives while maintaining coherence, improving performance across all metrics. Please refer to Appendix~\ref{exp:other_baseline} for additional streamlined human evaluations of the persona ablation and an example of the detailed critique process.
\begin{table}[t]
    \small
    \begin{tabular}{m{0.95\linewidth}}
    \toprule
    \multicolumn{1}{l}{\textbf{Without Persona Critiques (Plain Critiques)}} \\ 
    \midrule
    1. Question: How can the \negative{narrative structure} be altered to include multiple perspectives or overlapping ...\\
    \\
    2. Question: What if the \negative{storyline is restructured} to alternate between different character perspectives with each chapter, giving a multi-dimensional view ...\\
    \\
    3. Question: What if the invention not only reshapes reality but also \negative{unlocks an alternate dimension or parallel universe}, leading to a clash ...\\
    \midrule
    \multicolumn{1}{l}{\textbf{With Persona Critiques (Detailed Critiques)}} \\ 
    \midrule
    1. Question: How can the \plan{ethical dilemmas surrounding Jameson and David's research} extend beyond the scientific community and impact society ...\\
    \\
    2. Question: How does the advancement of technology in this futuristic world \plan{impact other aspects of society} beyond the field of medicine? ...\\
    \\
    3. Question: How does \plan{Cassie Olsen's personal connection} to one of the kidnapped subjects influence ...\\
    \bottomrule
    \end{tabular}
    \caption{Example of story plan comparison between without-persona-critique and with-persona critiques in \RePlan{}. Without personas, critiques are often irrelevant to the original storyline and are general.}
    \label{anlys:persona}
    \vspace{-17pt}
\end{table}
\subsection{Number of Rounds.}
\begin{figure}[t]
    \centering
    \includegraphics[width=0.99\columnwidth]{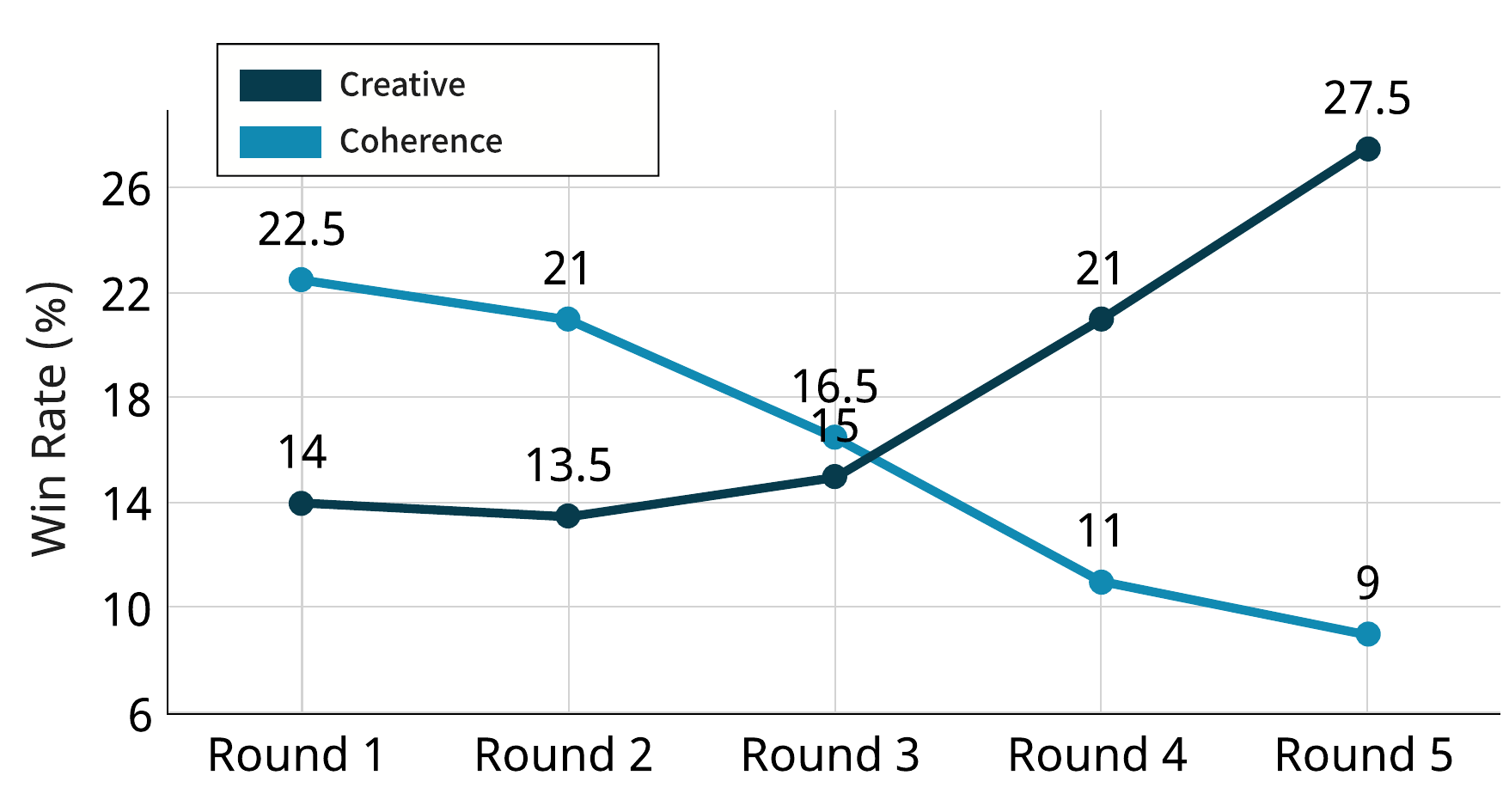}
    \caption{Pairwise comparison evaluations using GPT-4 (Win Rate \%) to analyze changes in story plan creativity and coherence with different critique iterations.}
    \label{fig:plot}
    \vspace{-10pt}
\end{figure} 
We investigate the impact of the number of revision rounds in the \RePlan{} on creativity and coherence within story plans. This involves evaluating 300 story plans, assessing their `Creativity' and `Coherence' across all rounds. A GPT-4 evaluator (detailed in Appendix~\ref{exp:autoamtic_anno}) is employed to determine which round is better than the others, or none if no preferable option exists.
The results, presented in Figure~\ref{fig:plot}, clearly demonstrate the trade-off between creativity and coherence. Due to this trade-off, identifying the optimal number of critique cycles for the best story plan becomes challenging. This underscores the importance of employing an evaluator in \RePlan{}, who examines story plans at each round to identify the optimal plan.
\subsection{Human-Machine Interactive Writing.}\label{abl:Iter}
\begin{figure}[t]
    \centering
    \includegraphics[width=0.99\columnwidth]{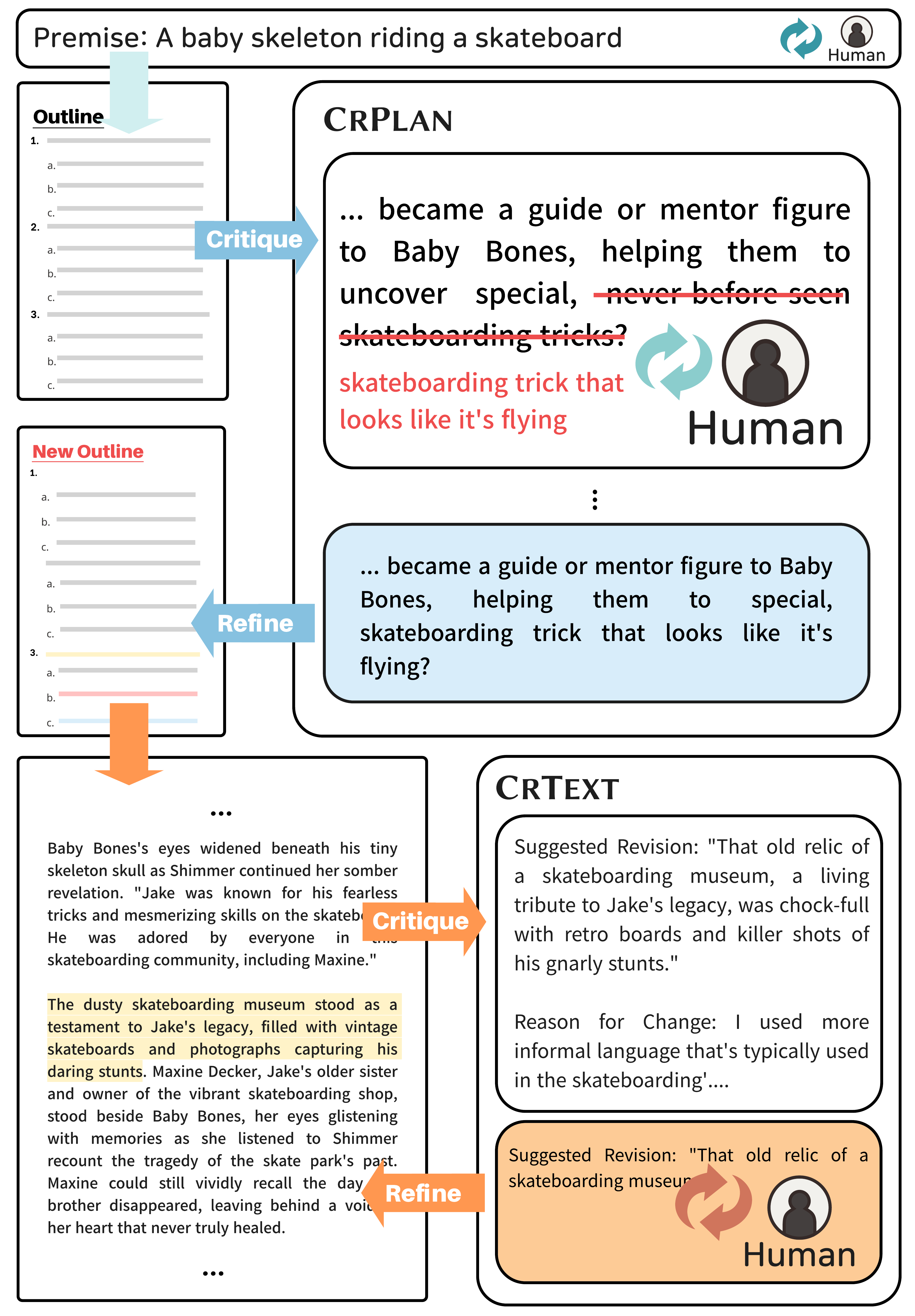}
    \caption{Machine-human interactive writing system allows human participation as any of the players in the revision process of \frameworkname{}.}
    \label{fig:machine-human}
\end{figure}
\frameworkname{} is designed to facilitate active participation from human writers in the revision process, enabling interactive collaboration between human and machine in creative writing. As illustrated in Figure~\ref{fig:machine-human},~\frameworkname{} enables human participants to either modify critiques generated by the system or write their own. Additionally, participants can take on the role of a leader, who is responsible for evaluating and selecting the critiques.

To substantiate the utility of our human-machine interactive system, we conduct a user experience experiment where the user took on the role of a leader module, evaluating effectiveness of the writing system. The experiments involve three annotators, each of whom generates ten stories, subsequently refines through three rounds. The detailed metrics are as follow:

\begin{itemize}
\itemindent=-10pt
    \item Edited: Whether the story changes due to the critique. This metric is marked as ``Pass'' if changes reflect the critique and ``Fail'' if not.
    \item Accepted: Whether the refined story aligns with the critique content provided by the user acting as the leader. This metric is marked as ``Pass'' if it aligns and ``Fail'' if it does not.
\end{itemize}
\begin{table}[t]
\centering
\resizebox{0.75\linewidth}{!}{
\begin{tabular}{l >{\centering\arraybackslash}m{0.38\linewidth}>{\centering\arraybackslash}m{0.12\linewidth}}
\hline
\textbf{Metrics} & \textbf{Pass Rate(\%)} & \textbf{Kappa}\\
\hline
Edited & 100.00 & 1.000 \\ 
Accepted & 83.33 & 0.3681 \\
\hline
\end{tabular}
}
\caption{Results of the user experience experiment with three annotators, each generating ten stories, totaling 30 user-led evaluations. Most inter-agreements are fair.}
\vspace{-12pt}
\label{tbl:user_ex}
\end{table}
As shown in Table~\ref{tbl:user_ex}, while the story changed in every turn due to the critique compared to the initial story, we find that the detailed content of the critiques does not always completely align with the changes. Detailed examples of story plans and extended narratives developed through this human-machine collaborative writing process, and screenshots of the web implementation, are provided in Appendix~\ref{example:all}.
\section{Conclusion}
We introduce \frameworkname{}, a novel framework that enhances story creativity and ensures narrative coherence through collaborative LLM critiques. The critics systematically critique the story plans (\RePlan) and text expression (\ReSent), employing diverse creative criteria. Their adaptive personas play a crucial role in maintaining narrative coherence, providing detailed critiques that help create a consistent, contradiction-free, and creative story.
Extensive human evaluation demonstrates that the \frameworkname{} significantly enhances the creative dimensions of stories while preserving narrative coherence. Also, \frameworkname{} is designed to facilitate interactive collaboration between humans and machines, allowing active participation of human writers in the critique process. We hope \frameworkname{} opens new dimensions in computational creativity, enabling the generation of unique and captivating narratives.
\section*{Limitation}
The \frameworkname{} framework refines narratives through critiques generated by LLMs, with its effectiveness closely tied to the capabilities of the specific models employed. In this study, we employ the DOC pipeline for story generation, which possibly limits the output format. However, the flexible working mechanism of our \frameworkname{} facilitates the exploration of alternative narrative planning pipelines.
In its current design, \frameworkname{} is optimally configured for English language applications. Adapting it for additional languages would require comprehensive revisions to the existing prompt structures. Additionally, in this study, the criteria we use are limited to those defined by \citet{mozaffari2013analytical}. However, it is also feasible to integrate other well-defined storytelling-related criteria into our mechanism..
\section*{Ethical Consideration}
Story refinement system of \frameworkname{} has the potential to transform existing stories in creative ways, but also risks introducing toxicity or falsehoods in the text, potentially causing harm. In this study, we have not attempt to filter out harmful text; however, \frameworkname{} is modularly constructed with respect to the language models used, so it can, in principle, adhere to the toxicity guidelines of these models. Similar to how \leaderGPT{} and \evalGPT{} is used in this tudy to create optimal critiques and stories, the integration of a toxic filtering module within the critique process could potentially reduce toxicity risks.
\section*{Acknowledgements}
We thank the reviewers for their valuable feedback and the entire \textit{Language \& Intelligence Lab} family for their helpful discussions. This work was supported by Institute of Information \& communications Technology Planning \& Evaluation(IITP) grant funded by the Korea government(MSIT)(No.RS-2020-II201336, Artificial Intelligence graduate school support(UNIST)) and the Leading Generative AI Human Resources Development(IITP-2024-RS-2024-00360227) grant funded by the Korea government(MSIT) and the 2022 Research Fund (1.220140.01) of UNIST(Ulsan National Institute of Science \& Technology).
\bibliography{anthology,custom}
\section*{Appendix}
\appendix
\section{Analysis Example}
\subsection{Adaptive Personas}
\label{abl:persona_adpative}
\begin{table*}[t]
\small
\begin{tabular}{m{0.99\linewidth}}
\toprule
\multicolumn{1}{l}{\textbf{Initial Plan}} \\
\midrule
\midrule
1. Calvin Blake meets Eleinor Grays and discovers that, unlike him, she is unable to feel. \\
\quad   a. Calvin Blake notices that Eleinor Grays struggles with \textbf{suppressing her emotions}, and realizes there may be something different about her. \\
\quad   b. Calvin tries to act naturally around her, not wanting to risk their cover, but also drawn to her lack of emotion. \\
\quad   c. Calvin and Eleinor are caught in a moment of feeling by James Blake. \\
2. Calvin and Eleinor must navigate the \textbf{societal constraints of a dystopian future} where emotions are outlawed, to be together and feel alive again. \\
 ...\\

\toprule
\multicolumn{1}{l}{\textbf{Adaptive Personas}} \\
\midrule
\midrule
Expert 1.\\
\quad    Profession: \textbf{Sociologist}\\
\quad    Feedback Focus:Societal dynamics and world-building\\
\quad    Feedback Focus Details: Expert 1 will focus on evaluating the societal constraints and dynamics within the dystopian world, such as the laws and regulations surrounding emotions. They will provide insights on whether the world-building is consistent and believable, and suggest ways to enhance the portrayal of the dystopian society. \\
\\
Expert 2.\\

\quad    Profession: \textbf{Psychologist specializing in emotions}\\
    
\quad    Feedback Focus: Emotional depth and authenticity\\
    
\quad    Feedback Focus Details: Expert 2, a psychologist specializing in emotions, can provide valuable insights into the emotional aspects of your story. They can evaluate the depth and authenticity of the emotions portrayed by your characters and suggest ways to enhance their emotional journeys for a more impactful and engaging storyline.\\
\\
Expert 3.\\

\quad    Profession: \textbf{Futurist} \\
    
\quad    Feedback Focus: Worldbuilding and believability\\
    
\quad    Feedback Focus Details: Expert 3, a futurist, can offer expertise in worldbuilding and ensuring the believability of your dystopian future. They can evaluate the consistency and coherence of your futuristic setting, as well as provide insights into possible societal developments and technological advancements that could enhance the overall plausibility of your novel.\\
\bottomrule
\end{tabular}
\caption{Example of adaptive personas related to the story elements like ``social, dystopian future, emotion''.}
\label{exam:adpative_persona}
\end{table*}
Table~\ref{exam:adpative_persona} presents examples of personas generated using the prompt template from Table~\ref{prompt:persona_creator}, adapted to the story's narrative. These personas are created in relation to the emotional turmoil of the character Elinor Grays, the dystopian future setting, and the societal constraints featured within the story. This allows for generating more detailed and relevant critiques of the narrative context.
\subsection{Other Criteria}
\label{abl:vairious_critieria}
We apply creative character and creative sentence structure criteria at each stage of our framework. Below are detailed explanations for each criterion:
\begin{itemize}
    \item Dynamic Development: Dynamic development refers to the evolution of a character over the story. This transformation is often driven by conflicts, challenges, and experiences that compel the character to change.
    \item Inverted and Non-linear Structures: Inverted structures involve rearranging the typical order of words or phrases to emphasize particular elements, whereas non-linear structures disrupt the chronological flow of the narrative, encouraging readers to piece together the timeline.
\end{itemize}
Table~\ref{exam:other_criteria} presents an example of applying dynamic development during the \RePlan{} phase. As illustrated in this example, events related to Gabriel's narrative loyalty to William emerge. This leads to a conflict situation for Gabriel, making the protagonist's character more dynamic.

Also, Table~\ref{exam:other_critieria_2} provides an example of applying inverted and non-linear structures during the \ReSent{} phase. In this example, by altering the typical subject-verb order to emphasize `serious danger' at the beginning of the sentence, the inverted sentence structure highlights the risks preceding the potential.
These examples demonstrate that our framework operates effectively across various creative criteria.
\begin{table*}[t]
\small
\begin{tabular}{m{0.99\linewidth}}
\toprule
\multicolumn{1}{l}{\textbf{Initial Plan}} \\
\midrule
\midrule
1. \negative{Gabriel, the loyal soldier, is disturbed by Jonas's unusual experiments and begins to distrust William.} \\
\quad    a.  Gabriel is initially loyal to William and defends his actions to Isabelle. Scene: the forest on the island, where Gabriel confronts Isabelle and Jared after they are caught snooping around.  \\
\quad    b. However, as Jonas's experiments yield frightening results, Gabriel starts to doubt William's motivations and fears for the safety of the island's inhabitants. Scene: the lab, where Gabriel confronts Jonas about his experiments. \\
\quad    c. He attempts to persuade his comrades to defy William's orders and stop Jonas. Scene: the barracks, where Gabriel speaks with the other soldiers about Jonas's experiments.  \\
...\\
\toprule
\multicolumn{1}{l}{\textbf{Critiques - Dynamic Development}}\\
\midrule
\midrule
Question: Could Gabriel Stone have a backstory where he owes his life or career to William Rhodes due to a \plan{critical incident in the past, thus complicating his loyalty?} How does Gabriel’s realization of the destructive nature of William’s experiments conflict with this debt, and what specific event triggers his decision to oppose William?\\
\toprule
\multicolumn{1}{l}{\textbf{Refined Plan}} \\
\midrule
\midrule
1. \plan{Gabriel's deep-seated loyalty is rooted in his past with William.}\\

\quad    a. Gabriel owes his life and career to William after a critical rescue during a military operation.\\
    
\quad    b. William mentors Gabriel, seeing potential in his resilience and moral compass. \\

2. \negative{Gabriel, now a loyal soldier, starts to question the morality of William's actions due to Jonas's unsettling experiments.}\\

\quad    a. Gabriel defends William's intentions to Isabelle but harbors doubts. Scene: The forest on the island, during a tense confrontation with Isabelle and Jared. \\
    
\quad    b. As experiments escalate, Gabriel's fear for the island's safety grows, conflicting with his indebtedness to William. Scene: The lab, during a heated discussion with Jonas.\\
    
\quad    c. Gabriel decides to oppose William after witnessing a particularly horrific experiment.\\...\\
\quad     b. \plan{Gabriel, conflicted by loyalty and morality, decides to protect Victoria and confront William.} Scene: The forest, as chaos ensues from the lab. Characters: Victoria Heller, William Rhodes, Gabriel Stone \\...\\
\bottomrule
\end{tabular}
\caption{Example of refining a story plan in the \RePlan{} stage using the `Dynamic Development' criterion.}
\label{exam:other_criteria}
\end{table*}
\begin{table*}[t]
\small
\begin{tabular}{m{0.99\linewidth}}
\toprule
\multicolumn{1}{l}{\textbf{Initial Text}} \\
\midrule
\midrule
...\\
Victoria's heart pounded in her chest as she took in Jonas's words. She couldn't deny the allure of this extraordinary power that he described. It was as if the lightning in those bottles held the key to unlimited possibilities, a force that could reshape the world. \negative{But along with that boundless potential came grave danger, and Victoria couldn't shake off her growing unease.}
Leaning forward, Victoria studied the intricate patterns of electricity dancing within the glass container. The blue tendrils writhed and twisted, illuminating Jonas's face with an otherworldly glow. It was mesmerizing, seductive even, but Victoria knew she couldn't let herself be swayed by its enchantment.\\
...\\
\toprule
\multicolumn{1}{l}{\textbf{Critiques - Inverted and Non-linear Structures}}\\
\midrule
\midrule
Original Sentence: But along with that boundless potential came grave danger, and Victoria couldn't shake off her growing unease.\\
\\
Suggested Revision: \textCRIT{Grave danger came with that boundless potential, and Victoria's growing unease remained unshaken.}\\
\\
Reason for Change: This revision uses an anastrophe by shifting the typical subject-verb order to emphasize ``grave danger'' right at the beginning, thereby highlighting the peril before the potential. The change also makes Victoria's emotional state more prominent, creating a stronger sense of tension and focus on her personal struggle.\\
\toprule
\multicolumn{1}{l}{\textbf{Refined Text}} \\
\midrule
\midrule
...\\
Victoria's heart pounded in her chest as she took in Jonas's words. She couldn't deny the allure of this extraordinary power that he described. It was as if the lightning in those bottles held the key to unlimited possibilities, a force that could reshape the world. \textCRIT{Grave danger came with that boundless potential, and Victoria's growing unease remained unshaken.}
Leaning forward, Victoria studied the intricate patterns of electricity dancing within the glass container. The blue tendrils writhed and twisted, illuminating Jonas's face with an otherworldly glow. It was mesmerizing, seductive even, but Victoria knew she couldn't let herself be swayed by its enchantment.\\
...\\
\bottomrule
\end{tabular}
\caption{Example of refining a story plan in the \ReSent{} stage using the `Inverted and Non-linear Structure' criterion.}
\label{exam:other_critieria_2}
\end{table*}
\subsection{Ablation Study Example}
\label{abl:example}
In \frameworkname{}, we incorporate a diverse array of LLM critics and modules. To assess the effectiveness of each module and the value of collaborative critique, we conduct three ablation experiments: Leader, Criteria,\footnote{Refining story using only single criterion.} and Persona. Our ablation study illustrates the impact of each module's removal with specific examples.
\paragraph{Role of Leader.}
\label{ablappend:leader}
Table~\ref{exam:leaderGPT} presents examples that compare story plans refined with leader involvement against those generated without it in \RePlan{}. In the example without a leader, the narrative faces criticism for handling the plotline of Jake's disappearance and the group's delve into political secrets simultaneously, leading to a noticeable inconsistency. Jake, previously noted as missing, unexpectedly returns in the subsequent scene with no given rationale. On the other hand, the example featuring a leader critique meticulously preserves the narrative's intricate layers. It carefully navigates through Jake's disappearance and explores the nuances of Sarah's relationship, thereby maintaining the story's seamless and logical progression. 

Table \ref{exam:leaderGPT_TEXT} illustrates the outcomes of applying all critiques without selecting any through \leaderGPT{} during the \ReSent{}. The example underscores the importance of tailoring text expressions or scenarios, such as character dialogues, with suitable creative criteria—for instance, incorporating colloquialisms like "y’know" to enhance authenticity. This shows the necessity for a system like \leaderGPT{}, which selects critiques based on the context, ensuring that refinements are appropriately applied to enhance the text's expression.
\begin{table*}[t]
\small
\begin{tabular}{m{0.99\linewidth}}
\toprule
\multicolumn{1}{l}{\textbf{Initial Plan}} \\
\midrule
\midrule
1. Jake and his friends are trying to stay one step ahead of their rivals, deciphering cryptic clues, and outwitting their enemies. \\
\quad a. Jake and his friends investigate a series of cryptic clues that lead them deep into the heart of the city. \\
\quad    b. They encounter dangerous obstacles, including security systems, guards, and rival treasure hunters who will stop at nothing to get their hands on the diamond. \\
\quad    c. The group narrowly escapes several close calls, relying on their wits and cunning to stay one step ahead of their enemies. \\
2. The stakes are high as they get closer to unraveling the grand heist, facing dangerous obstacles, and dealing with unexpected twists and turns. \\

\quad    a. As Jake, Sarah, Tom, Emma, and Mike get closer to uncovering the mystery of the treasure hunt, they begin receiving threatening messages from unknown sources. \\...\\

3. Amidst the chaos, they uncover a shocking conspiracy that threatens to change the city's political landscape and could put their own lives in jeopardy.\\...\\
\toprule
\multicolumn{1}{l}{\textbf{With Leader Critics}}\\
\midrule
\midrule
1. Jake, Sarah, and Tom form a team of adventure-seekers, aiming to uncover a significant city treasure before their rivals do. \\
\quad	a. Jake, the group's leader, has deep feelings for Sarah, but Sarah is \plan{romantically caught between Jake and Tom}, causing tension amongst them.\\
	
\quad    b. Despite their romantic complications, the trio, along with their friends Emma and Mike, embark on their treasure hunt, interpreting cryptic clues leading them to the heart of the city.\\
	
2. Encountering dangerous obstacles and having their rivalry caught inbetween city's warring factions, the group endures, using their wits and cunning to keep steps ahead.\\

\quad	a. As they get closer to the treasure, they face life-threatening challenges, from security systems and guards to rival treasure hunters.\\ 
	
\quad    b. Sarah's emotions for both Jake and Tom cause frictions within the team, adding to their difficulties. A rival faction, aware of their emotional turmoil, tries to exploit this as a weakness. \\

3. \plan{Jake gets captured} and held hostage at the city museum, causing panic within the team.\\

\quad	a. In Jake's absence, Tom shines as interim team leader, further escalating tensions between him and Jake.\\
	
\quad    b. Sarah, caught in between her feelings for Jake and her admiration for Tom, struggles to keep the team focused on their mission. \\

4. Amidst Jake’s captivity and their own internal challenges, they uncover a shocking political conspiracy that threatens the city's balance.\\

\quad	a. While working on a rescue mission for Jake, the group stumbles upon a conspiracy that involves the city's mayor and the treasure they are trying to find. \\
	
\quad    b. Sarah and Tom, fueled by their need to \plan{save Jake} and to uphold what's right for their city, put their feelings aside, setting the stage for an intense final showdown. ...\\
\toprule
\multicolumn{1}{l}{\textbf{Without Leader Critics}} \\
\midrule
\midrule
1. Jake and his friends are drawn into a city-wide treasure hunt, marked by cryptic clues\\

\quad    a. The adventure kicks off with Jake and friends investigating a series of cryptic clues, leading them into the heart of the city.\\
    
\quad    b. The story then shifts to the mayor's perspective, revealing his motives and plans, and his ties to the dangerous obstacles Jake's team face, like security systems and guards.\\
    
\quad    c. A rival group of treasure hunters chasing the same diamond come into focus, their relentless and ruthless pursuit throwing another roadblock in Jake's way. \\

2. As their quest intensifies, Jake's team grapples with danger and betrayal, while the city's political corruption begins to surface.\\

\quad    a. Jake, Sarah, Tom, Emma, and Mike draw closer to the truth, only to fall into a web of threats from unknown sources. The story, now narrated from their rivals' perspective, presents their shared ambition for the diamond.\\
    
\quad    b. As Jake gets abducted and is kept inside the museum, one of his friends is revealed as the traitor. The shocking twist adds a layer of emotional impact and conflict to the storyline.\\
    
\quad    c. Despite the treachery, the \negative{remaining group pushes forward without Jake}, their resolve doubling.\\

3. As \negative{Jake and his friends} delve deeper into the conspiracy, the diamond heist is revealed to be emblematic of the city’s overwhelming political corruption, leading to an intense showdown.\\

\quad    a. Their investigations expose a vast conspiracy, which extends to the museum and several political figures in the city. The story offers an insight into the mayor's perspective once more, revealing his illicit involvement. \\
    
\quad    b. The diamond heist morphs into a metaphor for the city's corrupt political structure, the mayor and his greed reflecting the dark power dynamics at play.\\
    
\quad    c. An all-out confrontation at the city museum serves as the climax, where Jake's team, the mayor's henchmen, and Daniel Lee's soldiers collide. Their treasure hunt, thus, transforms into a fight against corruption in a bid to reclaim their city.\\
\bottomrule
\end{tabular}
\caption{Example of story plan comparison between Without-Leader-Critique and With-Leader Critiques in \RePlan{}.}
\label{exam:leaderGPT}
\end{table*}
\begin{table*}[t]
    \small
    \begin{tabular}{m{0.95\linewidth}}
    \toprule
    \multicolumn{1}{l}{\textbf{Initial Text}} \\
    \midrule
    \midrule
    ...\\
    But hearing someone else acknowledge that fire within him ignited a spark of hope and possibility. \textCRIT{"I never thought anyone would understand,"} Alex murmured, his voice laced with vulnerability. He glanced around the dimly lit room, its dusty corners illuminated by the faint glow of holographic advertisements flickering outside. \\...\\
    \midrule
    \multicolumn{1}{l}{\textbf{Critiques}} \\
    \midrule
    \midrule
    1. Original Sentence: "I never thought anyone would understand" \\
    \\
    Suggested Revision:\plan{"Never did I reckon anyone would get it, y'know?}\\
    \\
    Reason for Change: This revision incorporates informal language ("y'know") and uses "reckon" as an unusual word choice, adding a touch of old-fashioned or regional flair.\\ \\
    2. Original Sentence: I never thought anyone would understand \\ 
    \\
    Suggested Revision:\negative{"I always believed my thoughts were whispers, too faint for anyone to truly hear."}\\
    \\
    Reason for Change: This revision incorporates the creativity feature of "Hear" by using the metaphor of thoughts being like "whispers," which also enhances the "Insight" aspect by describing the internal perception of one's own thoughts.\\
    \midrule
    \multicolumn{1}{l}{\textbf{Refined Text}} \\
    \midrule
    \midrule
    (1)
    ... 
    But hearing someone else acknowledge that fire within him ignited a spark of hope and possibility.\plan{"Never did I reckon anyone would get it, y'know,"} Alex murmured, his voice laced with vulnerability. He glanced around the dimly lit room, its dusty corners illuminated by the faint glow of holographic advertisements flickering outside ...\\
    \\
    (2) 
    ...  
    But hearing someone else acknowledge that fire within him ignited a spark of hope and possibility.\negative{"I always believed my thoughts were whispers, too faint for anyone to truly hear.,"} Alex murmured, his voice laced with vulnerability. He glanced around the dimly lit room, its dusty corners illuminated by the faint glow of holographic advertisements flickering outside ...\\
    \bottomrule
    \end{tabular}
    \caption{Example when both critiques are applied without \leaderGPT{} in \ReSent{}. For specific text expressions like character dialogues, critiques aimed at enhancing the Voice are more suitable than those enhancing the Image (e.g., colloquial expressions such as "y’know" in the example).}
    \label{exam:leaderGPT_TEXT}
\end{table*}
\begin{table*}[t]
\small
\begin{tabular}{m{0.95\linewidth}}
\toprule
\multicolumn{1}{l}{\textbf{Initial Story Plan}}\\
\midrule
\midrule
1. John \textCRIT{discovers the package of cash} and visits Abe every day to share his lunch, and they develop a bond.\\
\quad    a. John discovers the package of cash one day while walking his dog Max in the forest behind his house. \\

\quad    b. John visits Abe every day to share his lunch, and they develop a bond. \\

\quad    c. Abe decides to return the favor and repay John for his kindness. \\

2. With Abe's guidance, John starts an art therapy group at school, where he can connect with his peers and express himself creatively.\\

\quad    a. Abe guides John to start an art therapy group at school as a way to connect with his peers and express himself creatively.\\

\quad    b. John struggles to convince his classmates and school staff to allow the art therapy group to form, but eventually gains their support after sharing his own struggles with mental health. \\

\quad    c. John is able to find success and fulfillment as he becomes less anxious and more confident in himself. \\

3. While John's mental health improves, he becomes closer to Alyssa and Jake, who support him in exploring his artistic side and his aspirations. \\

\quad    a. John forms a growing friendship with Alyssa through their connection over mental health issues. \\...\\
\midrule
\multicolumn{1}{l}{\textbf{Refined Plan: `Originality Plot/Setting/Themes’ Criteria}}\\
\midrule
\midrule
1. John discovers the package of cash and wrestles with the \negative{ethical dilemma}, even as he grows closer to Abe.\\
\quad    a. John stumbles upon a package of cash in the forest behind his house while walking with his dog, Max. He's now faced with the ethical dilemma of keeping it or turning it in.\\
\quad    b. During this turmoil, John continues his visits to Abe, sharing his lunches with him and formulating a strong bond.\\
2. Learning about the \negative{digital world}, John integrates technology with art therapy and starts a group in his school, forming meaningful connections.\\

\quad    a. With Abe's guidance, John explores the potential of integrating digital media tech into art, creating a more interactive and expressive form of therapeutic art.\\
  
\quad    b. Despite the initial struggle to convince his school and classmates, John succeeds in establishing the digital-art therapy group, illustrating his personal mental health struggles and how this modernised form of art therapy might help.  \\
  
\quad    c. The success of the project, coupled with the fulfillment of helping his peers, alleviates John's anxiety and boost his confidence.  \\

3. John's renewed connection with Alyssa and Jake strengthens amidst the success of his \negative{digital-art therapy group}.  

\quad    a. By sharing his journey with mental health issues, John deepens his friendship with Alyssa, who shares a similar struggle.\\
  
\quad    b. Alyssa's support helps John explore his artistic ambitions as he navigates through discovering his new identity and purpose.  \\
  
\quad    c. This modernized art therapy group not only brings John closer to his classmates but also to Jake, a fellow artist with an affinity for the digital medium. This growing connection fosters further creativity and innovation within the group.\\
\midrule
\multicolumn{1}{l}{\textbf{Refined Plan : Three Criteria (Flashback Structure)}}\\
\midrule
\midrule
1. John is now leading a successful \plan{art therapy group} at the school, where he, along with others, express themselves creatively.\\

\quad	a. John struggles to convince his classmates and the school staff to allow the art therapy group to form, in flashbacks, we learn his persistent efforts. \\

\quad	b. After sharing personal struggles with his mental health, John gains their support that marked the beginning of the group.\\

2. With the group's foundation, John's mental health improves and his relationships with his peers Alyssa and Jake strengthen. They support his artistry and future dreams.\\

\quad	a. Through a flashback, we see the development of John's deep friendship with Alyssa, both connecting over mental health issues.  \\

\quad	b. With the influence of the therapy group, John starts to spend more time with Alyssa and Jake, another budding artist, as he navigates his renewed sense of self and purpose.\\

3. \plan{Flashbacks reveal John's past}, depicting \plan{his discovery of a package of cash} one day while walking his dog Max in the forest behind his house.\\

\quad	a. Intertwining the present with the past, the story shows how John started visiting Abe regularly, sharing his lunch, and forming a bond.\\

\quad	b. The benevolent Abe returns John's favor by guiding him to start the school's art therapy group, allowing John to connect with his peers and express himself differently.\\

4. The art therapy group's growth sees John becoming less anxious and more confident in himself, paralleling his journey of growth with the group's evolution. \\

\quad	a. Flashbacks intermittently portray the growth of the group alongside John's boosted confidence and reduced anxiety. \\
...\\
\bottomrule
\end{tabular}
\caption{Example of story plan comparison between Single-Criteria-Critique and Three-Criteria-Critiques in \RePlan{}. Full refined story plans is provided in Appendix~\ref{example:all}.}
\label{abl:one_critic_1}
\end{table*}
\begin{table*}[t]
\small
\begin{tabular}{m{0.95\linewidth}}
\toprule
\multicolumn{1}{l}{\textbf{Original Text}}\\
\midrule
\midrule
... Karen's heart raced as she processed their words. \textCRIT{"But... why me?”}she managed to stammer. "Throughout the centuries," the beings explained, "we have searched for someone with a pure heart and a deep connection to their surroundings. Someone who possesses an unwavering sense of empathy and compassion. You are that person, Karen." ...\\
\midrule
\multicolumn{1}{l}{\textbf{Refined Text: Single Criteria (`Image')}}\\
\midrule
\midrule
...Karen's heart raced as she processed their words. \negative{"But... why have I been chosen for this?”}she managed to stammer. "Throughout the centuries," the beings explained, "we have searched for someone with a pure heart and a deep connection to their surroundings. Someone who possesses an unwavering sense of empathy and compassion. You are that person, Karen.”...\\
\midrule
\multicolumn{1}{l}{\textbf{Refined Text: Two Criteria}}\\
\midrule
\midrule
\\...Karen's heart raced as she processed their words. \plan{"Umm, but... why me, exactly?"}she managed to stammer. "Throughout the centuries," the beings explained, "we have searched for someone with a pure heart and a deep connection to their surroundings. Someone who possesses an unwavering sense of empathy and compassion. You are that person, Karen."\\
\bottomrule
\end{tabular}
\caption{Example of critiques applied from `Image' only in \ReSent{}.}
\label{abl:one_critic_2}
\end{table*}

\paragraph{Diversity of Criteria.}
\label{ablappend:one critic}
Table~\ref{abl:one_critic_1} shows examples comparing story plans refined through a single criterion critique ('Originality in plot/setting/themes') with those refined using three criteria critiques within \RePlan{}. The single-criterion critique, which concentrates exclusively on originality, typically makes minor adjustments to particular story elements like ethical dilemmas or digital settings without significantly transforming the story structure. In contrast, the three-criteria critique approach prompts a more substantial alteration in the narrative's structure. This approach introduces fresh themes, exemplified by the emergence of an art therapy group storyline. It integrates narrative techniques such as flashbacks, leading to notable modifications in the story's framework and broadening the array of narrative elements.

Table \ref{abl:one_critic_2} presents examples of critiques from a single LLM based on the 'Image' criterion, which is constrained and cannot accommodate unique expressions tailored for special situations, such as conversational dialogue. As a result, relying solely on a single criterion for critiques can restrict the breadth of feedback, potentially leading to an undue emphasis on refining specific aspects of the text without adequately addressing nuances like natural conversational expressions.
\paragraph{Role of Persona.}
\label{ablappend:non-persona}
Table \ref{exma:non_persona} illustrates the contrast between persona and non-persona critiques. The examples reveal that non-persona critiques often suggest general story improvement, such as enhancing narrative structure or reorganizing the storyline. Additionally, they may produce critiques on themes disconnected from the main narrative, like suggestions about alternate dimensions or parallel universes that bear no relation to a story focused on new energy development. These broad and unrelated critiques can undermine the story's coherence and limit the diversity of constructive feedback.
\begin{table*}[t]
    \small
    \begin{tabular}{m{0.95\linewidth}}
    \toprule
    \multicolumn{1}{l}{\textbf{Original Plan}} \\ 
    \midrule
    \midrule
    1. Ethan Grey discovers a mysterious and powerful energy source that can reshape the fabric of reality and grants immense, otherworldly powers. \\
    \\
\quad    a. Ethan Grey discovers a mysterious and powerful energy source in a laboratory accident. \\
\quad    b. He successfully replicates the energy source, convinced that he can use it to make the world a better place. \\
\quad    c. His partners, Jameson Rhodes and Lily Grey, express skepticism and fear of the unexplained forces in the power source. \\
\\
2. Lily Grey and Zoe Grey are brought into the project, with Lily providing expertise in quantum physics and Zoe becoming an essential part of the team, navigating a world of complex technology and adult choices beyond her years. \\

\quad    a. Ethan Grey starts experimenting with the energy source, discovering various extraordinary applications, unaware of the hidden mysteries.\\
\quad    b. He invites his long-time research partner, Lily Grey, to investigate and theorize about the energy source's capabilities and potential dangers. \\
\quad    c. Jameson Rhodes disapproves of Ethan and Lily's experimentation and warns against their reckless abandonment of conventional laws of physics. \\
\\
3. The team realizes that the energy source holds a sinister secret, connected to a larger conspiracy involving tech billionaires like Thomas Weston and shadowy government agents like Ava Mays. \\
    \\
\quad    a. The team's research begins to take a dark turn when they learn more about the source's power.\\
\quad    b. They discover that the energy source could be connected to a larger conspiracy involving powerful figures like billionaire Thomas Weston and ex-government agent Ava Mays. \\
\quad    c. The team faces opposition from shady interests in the form of government agencies and corrupt corporations, each vying for control. \\
    \midrule
    \multicolumn{1}{l}{\textbf{Non-Persona Critiques List}} \\ 
    \midrule
    \midrule
    1. Question: How can the \negative{narrative structure} be altered to include multiple perspectives or overlapping timelines, allowing the readers to piece together the story from different angles?\\
    \\
    2. Question: What if the \negative{storyline is restructured} to alternate between different character perspectives with each chapter, giving a multi-dimensional view of the events unfolding?\\
    \\
    3. Question: What if the invention not only reshapes reality but also \negative{unlocks an alternate dimension or parallel universe}, leading to a clash between different versions of the characters and amplifying the stakes and conflicts they face?\\
    \midrule
    \multicolumn{1}{l}{\textbf{Persona Critiques List}} \\ 
    \midrule
    \midrule
    1. Question: How can the \plan{ethical dilemmas surrounding Jameson and David's research} extend beyond the scientific community and impact society at large in the futuristic world?\\
    \\
    2. Question: How does the advancement of technology in this futuristic world \plan{impact other aspects of society} beyond the field of medicine? For example, how has it influenced the economy, politics, and everyday life?\\
    \\
    3. Question: How does \plan{Cassie Olsen's personal connection} to one of the kidnapped subjects influence her investigation and the way she uncovers the truth behind the experiments?\\
    \bottomrule
    \end{tabular}
    \caption{Example of comparing non-persona critiques with persona critiques. Non-persona critiques, in comparison to persona critiques, are broader (e.g., narrative structure, storyline is reconstructed...) and create critiques that do not align with the storyline (e.g., critiques suggesting an out-of-place alternate dimension or parallel universe).}
    \label{exma:non_persona}
\end{table*}
\section{Evaluation Details}
\label{exp:eval_detail}
\subsection{Human Evaluation Details.}
\label{exp:humanevla_anno}
We hire a professional annotator agency to conduct a human evaluation. The questionnaire used is as in Table~\ref{tab:human eval} and~\ref{tab:human eval2}. We conduct a human evaluation with four annotators and verified statistical significance using the t-test. 
\begin{table}[t]
    \small
    \begin{tabular}{m{0.95\linewidth}}
    \toprule
    We are conducting a survey comparing two storylines with the same premise. The stories will be evaluated based on three key aspects:\\
\\
1. Interesting: The storyline's ability to engage and captivate the reader.\\
2. Coherence: The logical and seamless interlinking of narrative elements such as plot, characters, and themes, ensuring the story progresses understandably and compellingly for the audience.\\
3. Creative: The originality and inventiveness of the storyline, offer a fresh perspective compared to typical narratives.\\
4. Closer to the premise: The narrative themes of the premise and the storyline are shared.\\
\\
You will be asked to evaluate two storylines according to the following criteria and answer four questions based on the provided key aspects.\\
\\
Question:\\
\\
1. Which storyline do you prefer/find more interesting overall?\\
\quad(1). Storyline A\\
\quad(2). Storyline B\\
\quad(3). Both are about equally good\\
\quad(4). Neither is good\\\\
2. Which story has a more coherent overarching storyline?\\
\quad(1) Storyline A\\
\quad(2) Storyline B\\
\quad(3) Both are about equally good\\
\quad(4) Neither is good\\\\
3. Which story has a more creative storyline? \\
\quad(1) Storyline A\\
\quad(2) Storyline B\\
\quad(3) Both are about equally good\\
\quad(4) Neither is good\\\\
4. Are both storylines close to the premise?\\
\quad(1) Storyline A is close to the premise\\
\quad(3) Storyline B is close to the premise \\
\quad(3) Both storylines A, and B are equally close to the premise \\
\quad(4) Neither is close to the premise\\
    \bottomrule
    \end{tabular}
    \caption{Human evaluation questionnaire of \RePlan{}.}
    \label{tab:human eval}
\end{table}
\begin{table}[t]
    \small
    \begin{tabular}{m{0.95\linewidth}}
    \toprule
    In this survey, you will participate in an experiment that compares the quality of sentence expressions in two stories, each based on a previously seen storyline. The focus will be on three specific sentences within the presented story, marked in "[START]sentence[END]" format.\\
    \\
Example) Brad was watching her as she read, taking in the sight of her as he sat there patiently waiting for his father to be ready for their meeting. [START]He had always found Karen to be attractive, even before his wife’s death, but as he looked at her now his admiration of her was heightened.[END] He knew that if he were to ever start dating Karen, things would be very awkward with her throughout the time that Shannon’s grave remained in front of his house. \\ 
\\
You will be asked to evaluate these sentences according to the following criteria and answer four questions based on the provided criteria.\\
1. Coherence: This involves examining the relatedness of the sentences to the context and their inter-sentence causal and temporal dependencies.\\
2. Writing Style Consistency: Evaluate whether the style of the given sentence is consistent with the overall context of the story.\\
3. Interesting: Assess if the expression in the sentences is rich and engaging, adding an interesting element to the story.\\
4. Creative: Unique or rich expression different from the typical story\\
\\
1. Which sentence has better Coherence?\\
\quad(1) Sentence A\\
\quad(2) Sentence B\\
\quad(3) Both are about equally good\\
\quad(4) Neither is good\\
\\
2. Which sentence has better Writing Style Consistency?\\
\quad(1) Sentence A\\
\quad(2) Sentence B\\
\quad(3) Both are about equally good\\
\quad(4) Neither is good\\
\\
3. Which sentence has better Interesting?\\
\quad(1) Sentence A\\
\quad(2) Sentence B\\
\quad(3) Both are about equally good\\
\quad(4) Neither is good\\
\\
4. Between these two sentences, which one exhibits greater creative expression?\\
\quad(1) Sentence A\\
\quad(2) Sentence B\\
\quad(3) Both are about equally good\\
\quad(4) Neither is good\\
    \bottomrule
    \end{tabular}
    \caption{Human evaluation questionnaire of \ReSent{}.}
    \label{tab:human eval2}
\end{table}
\subsection{Automatic Evaluation Details.}
\label{exp:autoamtic_anno}
In the ablation study, the automatic evaluation employed GPT-4 with a temperature setting of 0, using the GPT-4 automatic evaluation prompt from~\cite{wang-etal-2022-improved}.
It's important to note that the outcomes of GPT-4's automatic evaluation can be significantly affected by the order in which content is presented and may exhibit instability. To mitigate this, two-story plans is presented to GPT-4 in a random order for evaluation. The pairwise evaluation prompt is detailed in Table~\ref{prompt:automaticEval}.
\begin{table}[t]
    \small
    \begin{tabular}{m{0.95\linewidth}}
    \toprule
    Here are two storyline excerpts.\\
    You shouldn’t be concerned about the completeness of the plot.\\
    Storyline A:\\
    $\{\texttt{Storyline A}\}$\\
    \\
    Storyline B:\\
    $\{\texttt{Storyline B}\}$\\
    \\
    \\
    Answer the following question:\\
    1) Overall, which story do you prefer/find more interesting? \\A / B/ C \\
    2) Overall, which story has a more coherent overarching plot? \\A / B / C \\
    3) Overall, which story has a more creative plot? \\A / B / C \\
    4) Overall, Are both storylines closer to the premise? \\BY / OA / OB / BN / UN \\
     \\
    After providing your explanation, output your final verdict by strictly following this format:\\\\
    ”[[A]]” if storyline A is better, ”[[B]]” if storyline B is better, and ”[[C]]” for a tie or unable to determine.
    "[[BY]]" if storyline A,B are eqaully closer to premise, "[[OA]]" if only storyline A is close to the premise, "[[OB]]" if only stroline B is closer to the premise, "[[BN]]" if neither is close to the premise, "[[UN]]" if unable to determine.\\\\
    example)\\
    1:[[A]], 2:[[B]], 3:[[B]], 4:[[BY]]\\
    \bottomrule
    \end{tabular}
    \caption{Prompt for Automatic Evaluation.}
    \label{prompt:automaticEval}
\end{table}
\begin{table}[t]
\renewcommand{\arraystretch}{1.3}
\centering
\resizebox{0.999\linewidth}{!}{
    \begin{tabular}{@{}lcccc@{}}
    \toprule[1.3pt]
      & \multicolumn{4}{c}{\textbf{GPT-4 Automatic Evaluation - Plan Comparison}} \\
    \cline{2-5}
    \textbf{Model} & \textbf{Interesting\textsuperscript{\textuparrow}} & \textbf{Coherence\textsuperscript{\textuparrow}} & \textbf{Creative\textsuperscript{\textuparrow}} & \textbf{Relevant\textsuperscript{\textuparrow}} \\  
    \midrule
    DOC & 53.00 & 64.00 & 60.00 & 99.00 \\
    \RePlan{} & 81.67 & 69.33 & 80.00 & 98.66 \\ [0.5ex]
    \midrule 
     & \multicolumn{4}{c}{\textbf{GPT-4 Automatic Evaluation - Text Comparison}} \\
    \cline{2-5}
    \textbf{Model} & \textbf{Interesting\textsuperscript{\textuparrow}} & \textbf{Coherence\textsuperscript{\textuparrow}} & \textbf{Consistency\textsuperscript{\textuparrow}} & \textbf{Creative\textsuperscript{\textuparrow}} \\
    \midrule
    DOC & 41.33 & 62.00 & 78.00 & 46.00 \\
    \ReSent{} & 64.00 & 71.33 & 76.67 & 65.33 \\
    \midrule 
     & \multicolumn{4}{c}{\textbf{Inter agreement - Plan Comparison}} \\
    \cline{2-5}
    \textbf{Annotators} & \textbf{Interesting\textsuperscript{\textuparrow}} & \textbf{Coherence\textsuperscript{\textuparrow}} & \textbf{Creative\textsuperscript{\textuparrow}} & \textbf{Relevant\textsuperscript{\textuparrow}} \\  
    \midrule
    Annotator 1 & 0.2838 & 0.3759 & 0.4345 & 0.1215 \\
    Annotator 2 & 0.3173 & 0.3168 & 0.4626 & 0.1488 \\
    Annotator 3 & 0.3251 & 0.4021 & 0.3251 & 0.1623 \\
    \midrule 
     & \multicolumn{4}{c}{\textbf{Inter agreement - Text Comparison}} \\
    \cline{2-5}
    \textbf{Annotators} & \textbf{Interesting\textsuperscript{\textuparrow}} & \textbf{Coherence\textsuperscript{\textuparrow}} & \textbf{Consistency\textsuperscript{\textuparrow}} & \textbf{Creative\textsuperscript{\textuparrow}} \\
    \midrule					
    Annotator 1 & 0.2975 & 0.2698 & 0.3134 & 0.3529 \\
    Annotator 2 & 0.3728 & 0.3713 & 0.2338 & 0.3181 \\
    Annotator 3 & 0.1734 & 0.4359 & 0.6223 & 0.4945 \\
    \bottomrule[1.5pt]
    \end{tabular}
}
\caption{Results of the Cohen's kappa score evaluation, demonstrating a fair to moderate level of agreement among annotators, indicating the reliability of GPT-4's automatic evaluation.}
\label{tbl:rel_gpt}
\vspace{-15pt}
\end{table}
To verify the reliability of GPT-4's automatic evaluation and confirm its alignment with human judgments, we compare the human evaluation results from Section~\ref{exp:main} with the automatic evaluation results. In Section~\ref{exp:main}, three annotators evaluate 300 stories, and we calculate the Cohen's Kappa score between each annotator's evaluations and GPT-4's automatic evaluation of the same stories. As shown in Table~\ref{tbl:rel_gpt}, we measure each annotator's Cohen's kappa score and found a fair to moderate level of agreement, indicating the reliability of GPT-4's automatic evaluations. The results show that both human and GPT-4 evaluations exhibit similar trends with lower kappa scores for ``Relevant'' due to data bias, where most annotators rated options as `good', suggesting alignment between GPT-4's automatic evaluations and human judgments.
\label{appendix:example}
\section{Streamlined Human Evaluation}
\label{exp:other_baseline}
We conduct in-house human evaluations for the assessment of various backbones, baselines, and persoan ablation experiments. For each experiment, ten stories are assessed, with three annotators assigned to each story to evaluate inter-annotator agreement. All experiments demonstrate at least fair agreement among annotators.
\subsection{Re3 Comparison Experiments.}
As shown in Table~\ref{tbl:Re3}, \RePlan{} demonstrates superior performance in two metrics—Creativity and Coherence—by a large margin and scores slightly higher in Relevance. Additionally, \ReSent{} exhibits significantly higher performance in two metrics, Interestingness and Creativity, and also shows comparable performance in other metrics.
This demonstrates that the \frameworkname{} of the Re3 baseline also enhance creativity while maintaining the coherence of the story.
\begin{table}[t]
\renewcommand{\arraystretch}{1.3}
\centering
\resizebox{0.999\linewidth}{!}{
    \begin{tabular}{@{}lcccc@{}}
    \toprule[1.3pt]
      & \multicolumn{4}{c}{\textbf{Plan Comparison}} \\
    \cline{2-5}
    \textbf{Model} & \textbf{Interesting\textsuperscript{\textuparrow}} & \textbf{ Coherence\textsuperscript{\textuparrow}} & \textbf{Creative\textsuperscript{\textuparrow}} & \textbf{Relevant\textsuperscript{\textuparrow}} \\ 
    \midrule    
    Re3 & 65.00 & 70.00 & 10.00 & 75.00 \\
    \RePlan{} & 65.00 & 85.00 & 65.00 & 80.00 \\ [0.5ex]
    Fleiss' Kappa & 0.318 & 0.245 & 0.429 & 0.630\\
    \midrule
    & \multicolumn{4}{c}{\textbf{Text Comparision}} \\
    \cline{2-5}
    \textbf{Model} & \textbf{Interesting\textsuperscript{\textuparrow}} & \textbf{ Coherence\textsuperscript{\textuparrow}} & \textbf{Consistency\textsuperscript{\textuparrow}} & \textbf{Creative\textsuperscript{\textuparrow}} \\  
    \midrule    
    Re3 & 40.00 & 57.00 & 40.00 & 65.00 \\
    \ReSent{} & 60.00 & 55.00 & 60.00 & 78.30 \\ [0.5ex]
    Fleiss' Kappa & 0.204 & 0.309 & 0.271 & 0.297\\
    \bottomrule[1.5pt]
    \end{tabular}
}
\caption{Result of the in-house human evaluation of pairwise comparisons of~\frameworkname{} against the Re3 involved  comparing 10 storie. Most inter-agreements are fair to moderate.}
\label{tbl:Re3}
\end{table}
\subsection{GPT-4 Comparison Experiments.}
As shown in Table~\ref{tbl:GPT-4Doc}, \RePlan{} demonstrates superior performance on three metrics by a large margin and scores slightly higher in Relevance. Additionally, \ReSent{} exhibits significantly higher performance in two metrics, Interestingness and Creativity, and also shows comparable performance in other metrics.
This demonstrates that the \frameworkname{} of the GPT-4 backbone also enhance creativity while maintaining the coherence of the story.
\begin{table}[t]
\renewcommand{\arraystretch}{1.3}
\centering
\resizebox{0.999\linewidth}{!}{
    \begin{tabular}{@{}lcccc@{}}
    \toprule[1.3pt]
      & \multicolumn{4}{c}{\textbf{Plan Comparision}} \\
    \cline{2-5}
    \textbf{Model} & \textbf{Interesting\textsuperscript{\textuparrow}} & \textbf{ Coherence\textsuperscript{\textuparrow}} & \textbf{Creative\textsuperscript{\textuparrow}} & \textbf{Relevant\textsuperscript{\textuparrow}} \\
    \midrule    
    DOC (GPT-4) & 75.00 & 80.00 & 65.00 & 70.00 \\
    \RePlan{} & 80.00 & 95.00 & 100.00 & 80.00 \\ [0.5ex]
    Fleiss' Kappa & 0.333 & 0.315 & 0.378 & 	0.444\\
    \midrule
    & \multicolumn{4}{c}{\textbf{Text Comparision}} \\
    \cline{2-5}
    \textbf{Model} & \textbf{Interesting\textsuperscript{\textuparrow}} & \textbf{ Coherence\textsuperscript{\textuparrow}} & \textbf{Consistency\textsuperscript{\textuparrow}} & \textbf{Creative\textsuperscript{\textuparrow}} \\  
    \midrule    
    DOC (GPT-4) & 71.70  & 83.30 & 80.00 & 58.30 \\
    \ReSent{} & 91.70 & 96.60 & 66.70 & 93.30 \\ [0.5ex]
    Fleiss' Kappa & 0.293 & 0.323 & 0.267 & 0.312\\
    \bottomrule[1.5pt]
    \end{tabular}
}
\caption{Result of the human evaluation of pairwise comparisons of DOC (GPT-4) against the baseline involved  comparing 10 storie. Most inter-agreements are fair to moderate.}
\label{tbl:GPT-4Doc}
\end{table}
\subsection{Persona Effectiveness Experiments.}
\paragraph{Persona Ablation Experiments.}
The results, illustrated in Table~\ref{tbl:persona-abl}, demonstrate that personas contribute to fostering diverse critical perspectives while maintaining coherence, leading to performance improvements across all assessed metrics.
\begin{table}[t]
\renewcommand{\arraystretch}{1.3}
\centering
\resizebox{0.999\linewidth}{!}{
    \begin{tabular}{@{}lcccc@{}}
    \toprule[1.3pt]
      & \multicolumn{3}{c}{\textbf{Persona Ablation - Plan Comparison}} \\
    \cline{2-4}
    \textbf{} & \textbf{Interesting\textsuperscript{\textuparrow}} & \textbf{ Coherence\textsuperscript{\textuparrow}} & \textbf{Creative\textsuperscript{\textuparrow}} \\
    \midrule    
    Non-Persona & 80.00 & 70.00 & 60.00  \\
    Persona & 85.00 & 95.00 & 65.00  \\ [0.5ex]
    Fleiss' Kappa & 0.340 & 0.450 & 0.267 \\
    \bottomrule[1.5pt]
    \end{tabular}
}
\caption{Results of the human evaluation of persona ablation experiments involved comparing 10 story plans. Most inter-agreements are fair to moderate.}
\label{tbl:persona-abl}
\end{table}
\paragraph{Detailed Critique Process}
As shown in Table~\ref{exam:detailed_critique}, the detailed critiques produced in this manner help improve the story's coherence by allowing for precise modifications of the existing narrative content. In the initial story plan, there is only a brief mention that Shannon discovers the ugly realities of the inner city. However, the creation of a social worker persona within the inner city leads to the generation of critiques that integrate social issues into the narrative. This results in a more detailed depiction of the harsh realities faced by inner-city communities. Such detailed storytelling gives readers the impression that the story is thorough, with no missing content and a clear progression, leading to high marks for narrative coherence.
\begin{table*}[t]
    \small
    \begin{tabular}{m{0.95\linewidth}}
    \toprule
    \multicolumn{1}{l}{\textbf{Initial Plan}} \\
    \midrule
    \midrule
    ...\\
    Outline:\\
    \\
    1. Shannon's father, Mike, dies unexpectedly, leaving her determined to follow in his footsteps and become a successful journalist. Scene: Characters: Shannon Doyle, Mike Doyle\\
    \\
    \quad a. Shannon's father, Mike, dies unexpectedly. Scene: Characters: Shannon Doyle, Mike Doyle \\
    \\
    \quad b. Shannon decides to follow in her father's footsteps and become a successful journalist. Scene: Characters: Shannon Doyle, Mike Doyle\\ 
    \\
    2. Shannon lands her first major assignment, a feature on the inner city, but quickly discovers that the ugly reality of life in the city is far different from the dream she imagined. Scene: Characters: Shannon Doyle, Lena Saunders\\
    \\
    \quad a. Shannon lands her first major assignment, a feature on the inner city. Scene: Characters: Shannon Doyle, Lena Saunders \\
    \quad b. Shannon quickly \negative{discovers that the ugly reality of life in the city is far different from the dream she imagined.} Scene: the inner city. Characters: Shannon Doyle\\
    ...\\
    \midrule
    \multicolumn{1}{l}{\textbf{Adaptive Persona}} \\
    \midrule
    \midrule
    …\\
    Expert 2.\\
    \\
    \quad Profession: Inner City Social Worker\\

    \quad Feedback Focus: Inner city portrayal and social issues

    \quad Feedback Focus Details: Expert 2, an experienced inner city social worker, will evaluate the portrayal of the inner city and its social issues in the story. They will provide feedback on the accuracy and sensitivity of the representation, ensuring that the challenges faced by the characters reflect the realities of life in the inner city.\\
    …\\
    \midrule
    \multicolumn{1}{l}{\textbf{Critiques}} \\
    \midrule
    \midrule
    Question : How can Shannon's journey in the inner city lead to a deeper exploration of the power \plan{dynamics and social injustices} that exist within the community, ultimately leading to a transformative ending?\\
    \\
    Why: This question introduces a new layer of complexity to the storyline by focusing on power dynamics and social injustices within the inner city. It offers the potential for a transformative ending that challenges the status quo and adds depth to the narrative. The question aligns with the focus of an urban sociologist, as it encourages a realistic exploration of the inner city's social dynamics. The adjustments made based on this question have the potential to enhance the originality of the story, improve coherence, and create a more engaging and thought-provoking narrative.\\
    \midrule
    \multicolumn{1}{l}{\textbf{Refined Plan}} \\
    \midrule
    \midrule
    …\\
    2. The confrontation with corruption escalates, highlighting the personal and professional stakes for Shannon and Gary. Shannon's resolve, supported by informants who admired her father, leads to a critical showdown with the corrupt official, drawing on her father's courage and the support of the community. Scene: Characters: Shannon Doyle, Unnamed Corrupt Official, Unnamed Informants\\   
    \\
    \quad a. Shannon's determination, fueled by her father's legacy, faces the ultimate test against the corrupt official, reflecting the depth of her commitment to truth. Scene: Secluded city office Characters: Shannon Doyle, Unnamed Corrupt Official\\
    \\
    \quad b. The support from informants and the community bolsters Shannon's fight, leading to a pivotal moment of confrontation, echoing her father's resolve. Scene: Various city locations Characters: Shannon Doyle, Unnamed Informants\\
    ...\\
    \bottomrule
    \end{tabular}
    \caption{Example of an adaptive persona through a detailed critique process}
    \label{exam:detailed_critique}
\end{table*}
\section{Implementation Details}
We utilize the official DOC pipeline~\cite{doc_github} implementation provided by the DOC authors as our baseline to generate the initial plan and story. This involves employing a 2-layer initial plan and setting ChatGPT as the backbone for the DOC pipeline. The implementation details are as follows below.
\label{prompt:all}
\subsection{\RePlan{} Implementation Details.}
In \RePlan{}, we employ a method where leaders and critics utilize ChatGPT (GPT-3.5-turbo), set at a temperature of 1, to generate critiques of story plans. A persona creation guide is followed to ensure narrative coherence throughout this process, as outlined in . Critics are tasked with crafting three distinct critiques based on predefined criteria, from which they will choose one to explore in greater depth. This process is followed by specific prompts and criteria detailed in Table~\ref{prompt:persona_creator},~\ref{prompt:ending},~\ref{prompt:strucutre},~\ref{prompt:originality},~\ref{prompt:critique_creator} and~\ref{prompt:select_critiques}. 

The next step involves refining these story plans based on the critiques, guided by prompts in Table~\ref{prompt:refinement}. The evaluator sees the selection of one optimal revised story plan from the refined versions. This selection relies on a process adapted from~\citet{wang-etal-2022-improved}, which uses a GPT-4 automatic evaluation mechanism with a temperature setting of 0. To prevent the story plan sequence from influencing the evaluation outcome, we randomize their order, following best practices to minimize bias. The evaluators use a prompt for this task followed in Table~\ref{prompt:eval_gpt}.
\subsection{\ReSent{} Implementation Details.}
In \ReSent{}, critics and leaders have utilized ChatGPT, setting the temperature to 1, to generate critiques. For voice critiques, the prompts are aligned with the guidelines set out in Table~\ref{prompt:voice_critic}, whereas for image critiques, the prompts follow the directives in Table~\ref{prompt:image_critic}. The prompts used by leaders for their contributions are by the specifications laid out in Table~\ref{prompt:sentence_select}.

\begin{table}[t]
    \small
    \begin{tabular}{m{0.95\linewidth}}
    \toprule
    I have to improve the story plan of my novel. I need experts to give your current story plan a critical evaluation so I can develop it. These experts are relevant to the story plan I've presented. Create three persona for these experts, including their Profession, Feedback Focus Details, and Feedback Focus. Also, create a persona of a leader who checks the opinions of three experts and adopts the opinion of one. Following the below format.\\
    ---------------------------------\\
    Expert 1.\\
    Profession: // ... Profession ... // \\
    Feedback Focus:  // ... Expert 1's Feedback Focus ... // \\
    Feedback Focus Details: //... Expert 1's Feedback Focus Details... // \\
    Expert 2. \\
    Profession: // ... Profession ... // \\
    Feedback Focus:  // ... Expert 2's Feedback Focus ... // \\
    Feedback Focus Details: //... Expert 2's Feedback Focus Details... // \\
    Leader. \\
    Profession: // ... Profession ... // \\
    Feedback Focus:  // ... Leader's Feedback Focus ... // \\
    Feedback Focus Details: //... Leader's Feedback Focus Details... //\\
    ---------------------------------\\
    For reference, here is my story: $\{\texttt{story}\}$. The experts and leader should provide insights that help me deepen the narrative and develop the story further.\\
    \bottomrule
    \end{tabular}
    \caption{Prompt for persona creator.}
    \label{prompt:persona_creator}
\end{table}
\begin{table}[t]
    \small
    \begin{tabular}{m{0.95\linewidth}}
    \toprule
    (1) Unexpected Conclusions: This aspect includes sentences that wind up in an unusual or surprising way, challenging the reader's expectations set by the initial part of the sentence.\\ 
\\
(2) Humorous or Witty Conclusions: These are endings that incorporate humor or clever plays on words lending an element of surprise and entertainment. This feature contributes substantially to the overall unique voice of the writer. \\
\\
(3) Provocative or Intriguing Statements: This characteristic includes endings that are provocative or mysterious, prompting the reader to think deeper, question, and engage more with the content. \\
    \bottomrule
    \end{tabular}
    \caption{Prompts for "Unusual Ending" criteria.}
    \label{prompt:ending}
\end{table}
\begin{table}[t]
    \small
    \begin{tabular}{m{0.95\linewidth}}
    \toprule
    (1) Non-linear timeline: Stories do not have to unfold in a straightforward, chronological manner. Experiment with flashbacks, time skips, and non-linear timelines to make the narrative more unexpected. \\
\\
(2) Shifting perspectives: Altering the narrative perspective throughout the story can provide fresh insights and create intrigue. This can include alternating between first-person and third-person views, or switching between different characters' perspectives. \\
\\
(3) Intertextuality: Include references to other works, stories within stories, or use allegory as a structural device. This can create layers of meanings and associations that enrich the narrative. \\
\\
(4) Metafiction: Break the fourth wall by having characters acknowledge they're part of a story or by discussing elements of storytelling within the plot. This can create a self-aware story that directly engages with readers. \\
    \bottomrule
    \end{tabular}
    \caption{Prompts for "Unusual Story Structure" criteria.}
    \label{prompt:strucutre}
\end{table}
\begin{table}[t]
    \small
    \begin{tabular}{m{0.95\linewidth}}
    \toprule
    (1) Unconventional Themes: This category includes themes that are not typically encountered in everyday discourse. This could include themes from different cultures, underground societies, or niche hobbies and interests.\\
   \\
(2) Unique Plot: Succinct plots that deviate from standard, commonly seen narratives score higher in this category. This could involve unexpected plot twists, unconventional story progression, or atypical character development.\\
\\
(3) Diverse Settings: Diverse settings refer to the use of unfamiliar or striking locations and times - past, future, or entirely imaginative locations. These could range from sci-fi cityscapes, historical periods, to unique micro-settings such as a single room or a mystical forest.\\
\\
(4) Authenticity: This feature measures the realness of the theme/plot/setting for the reader. The use of vivid descriptions, consistent details, and emotionally engaging elements can contribute to a more authentic feel. \\
    \bottomrule
    \end{tabular}
    \caption{Prompts for "Original Theme/Plot/Setting" criteria.}
    \label{prompt:originality}
\end{table}
\begin{table}[t]
    \small
    \begin{tabular}{m{0.95\linewidth}}
    \toprule
    Look at my storyline above and make two requests\\
    1. First Request - \\Originality Questions for Storyline: You are seeking three questions that this storyline has "$\{\texttt{critic\_type\}}$". These questions should encourage thinking about unique elements or perspectives that can be added to the story. Remember to align your suggestions and critiques with your "professional background" and "expertise", focusing on aspects that would realistically occur or be relevant in your field.\\\\
    2. Second Request - \\Evaluation and Selection of the Best Question: Out of the three questions provided, you want to identify the best one that improves the originality of the story. This evaluation will be based on three factors:\\\\
    1) Originality: Does altering the storyline in response to this question enhance its originality by integrating "$\{\texttt{critic\_type\}}$" into the narrative?\\
    2) Coherence: Will adjustments made to the storyline based on this question improve its overall coherence and consistency?\\
    3) Interesting: Does refining the storyline according to this question amplify the storyline's appeal and keep the readers more engaged?\\\\
    The selected question will then be evaluated using these criteria.\\
    \bottomrule
    \end{tabular}
    \caption{Prompt for creating story plan's critic.}
    \label{prompt:critique_creator}
\end{table}
\begin{table}[t]
    \small
    \begin{tabular}{m{0.95\linewidth}}
    \toprule
        $<$3 Questions$>$\\
    1) $\{\texttt{first\_critique\}}$\\
    2) $\{\texttt{second\_critique\}}$\\
    3) $\{\texttt{third\_critique\}}$\\
    $<$Story Plan$>$\\
    $\{\texttt{story\_plan\}}$\\
    --------------------------\\
    $[$Request$]$\\ 
    My request is "Of the '3 questions' I posed, which is the best critique to improve the originality of the story plan? Please choose critiques based on the evaluation criteria below." \\
        1) Originality: Does altering the storyline in response to this question enhance its originality by integrating "original plot/setting/themes," "unusual story structure," and "unusual ending" into the narrative? \\
        2) Coherence: Will adjustments made to the storyline based on this question improve its overall coherence and consistency? \\ 
        3) Interesting: Does revising the storyline according to this question amplify the storyline's appeal and keep the readers more engaged? \\
    Choose one best question and ask the answer.\\
    \bottomrule
    \end{tabular}
    \caption{Prompt for \leaderGPT{} in \RePlan{}.}
    \label{prompt:select_critiques}
\end{table}
\begin{table}[t]
    \small
    \begin{tabular}{m{0.95\linewidth}}
    \toprule
    I have a storyline that I need to modify based on specific feedback from a critical review. The task involves integrating insights from the given critique into the existing storyline while adhering to certain constraints and format.\\
\\
Task:
Use the provided critical feedback to revise the given storyline. Ensure that the modifications align with the feedback's insights and maintain the original storyline's format.\\
\\
Constraints:
\\
1. Maintain the original format of the storyline as provided.\\
2. It is acceptable to change the order of the scenes as you see fit. \\
3. The outline must contain detailed descriptions of the events.\\
4. It is acceptable to add scenes as you see fit.\\ 
\\
Provided Materials:\\
\\
1.Critical Feedback: $\{\texttt{final critic}\}$\\
2.Original Storyline: $\{\texttt{story plan}\}$\\
    \bottomrule
    \end{tabular}
    \caption{Prompts for plan refinement.}
    \label{prompt:refinement}
\end{table}
\begin{table}[t]
    \small
    \begin{tabular}{m{0.95\linewidth}}
    \toprule
    Here are two story plan excerpts.
    You shouldn’t be concerned about the completeness of the plot.\\
    $\{\texttt{story\_set\}}$\\
    Task 1 question:\\
    1) Overall, which story do you prefer/find more interesting? A / B ... \\
    2) Overall, which story has a more coherent overarching plot? A / B ... \\
    3) Overall, Which story has a more creative plot? A / B ... \\ 
    4) Overall, which story’s plot is closer to the premise? A / B ... \\
    After providing your explanation, output your final verdict strictly below format:\\
    $\{\texttt{select\_generation\}}$ and $[[\texttt{TI}]]$ for a tie or unable to determine.\\
    \bottomrule
    \end{tabular}
    \caption{Prompt for \evalGPT{}.}
    \label{prompt:eval_gpt}
\end{table}
\begin{table}[t]
    \small
    \begin{tabular}{m{0.95\linewidth}}
    \toprule
    Sentence) \\
    $\{\texttt{text}\}$\\
    ===================\\
    Please review the following 'Sentence' from my draft and suggest revisions with explanations for each.\\
    However, when fixing a sentence, consider the following creativity features.\\
    \\
    Creativity Feature:\\ \\
    (1) Insight: This category contains words such as “think,” “know,” “consider”—words that can be used to describe thoughts, feelings, and internal images (“I imagined opening my arms and leaping off the balcony”).\\
    (2) See: This contains words such as “view” or “saw,” which can describe visual images.\\
    (3) Hear: This contains words such as “listen” and “hearing,” which are relevant to describe sound experiences.\\
    (4) Feel: This contains words, such as “feels” or “touch,” that can describe feelings and bodily sensations (e.g., “she feels a strange tingling sensation”). \\
    (5) Body: This contains words, such as “cheek” or “hands,” that are useful to describe feelings and bodily sensations (e.g., “My mouth was dry, and I felt my knees buckle.”).\\
    \bottomrule
    \end{tabular}
    \caption{Prompt for image critique.}
    \label{prompt:image_critic}
\end{table}
\begin{table}[t]
    \small
    \begin{tabular}{m{0.95\linewidth}}
    \toprule
    Text) \\
    $\{\texttt{text}\}$\\
    ===================\\
    Please review the following 'five sentences' from my draft and suggest revisions with explanations for each.\\
    However, when fixing a sentence, consider the following below creativity features.\\
    \\
    Creativity Feature:\\
    (1) Informal language: This category comprises informal language such as swear words, netspeak (“lol”), and nonfluencies like “er,” “umm,” relevant to the scoring of Voice.\\ 
    (2) Unusual words: Choice of particular or unusual words (e.g., rare or old-fashioned words or informal words\\
    (3) Noteworthy sentence structures: Number of words per sentence, Punctuation, and Use of commas specifically. \\
    (4) Authenticity: This variable measures how personal and honest a person's language sounds to listeners.
    \\
    \bottomrule
    \end{tabular}
    \caption{Prompt for voice critique.}
    \label{prompt:voice_critic}
\end{table}
\begin{table}[t]
    \small
    \begin{tabular}{m{0.95\linewidth}}
    \toprule
    I have a story that has undergone sentence refinements by a literary expert to enhance its creativity, considering specific 'Image' and 'Voice' creativity features. I need assistance in evaluating these changes to determine which are most effective in strengthening the narrative quality.\\\\
    Image creativity feature)\\
    (1) Insight: This category contains words such as “think,” “know,” “consider”—words that can be used to describe thoughts, feelings, and internal images (“I imagined opening my arms and leaping off the balcony”).\\
    (2) See: This contains words such as “view” or “saw,” which can describe visual images.\\
    (3) Hear: This contains words such as “listen” and “hearing,” which are relevant to describe sound experiences.\\
    (4) Feel: This contains words, such as “feels” or “touch,” that can describe feelings and bodily sensations (e.g., “she feels a strange tingling sensation”).\\
    (5) Body: This contains words, such as “cheek” or “hands,” that are useful to describe feelings and bodily sensations (e.g., “My mouth was dry, and I felt my knees buckle.”).\\
    \\\\
    Voice creativity feature)\\
    (1) Informal language: This category comprises informal language such as swear words, netspeak (“lol”), and nonfluencies like “er,” “umm,” relevant to the scoring of Voice. \\
    (2) Unusual words: Choice of particular or unusual words (e.g., rare or old-fashioned words or informal words\\
    (3) Noteworthy sentence structures: Number of words per sentence, Punctuation, and Use of commas specifically \\
    (4) Authenticity: This variable measures how personal and honest a person's language sounds to listeners.\\
    \\
    Task:\\
    From the list of 2 sentence refinements provided, select one sentence that most effectively enhance the narrative quality of the story. For each chosen refinement, provide a reason explaining why it strengthens the story.\\
    \\
    Sentence Refinements for Review:\\
    \\
    $<$Refinements set related to 'Image Creativity Feature'$>$\\
    $\{\texttt{Image refinement}\}$\\
    $<$Refinements set related to 'Voice Creativity Feature'$>$\\
    $\{\texttt{Voice refinement}\}$\\
    \\
    Please base your selections on the impact of these refinements on the story's overall creativity and narrative quality, considering the original story and the specified creativity features.\\
    \bottomrule
    \end{tabular}
    \caption{Prompt for \leaderGPT{} in \ReSent{}.}
    \label{prompt:sentence_select}
\end{table}
\section{Full Stories Example}
\label{example:all}
\subsection{Flashback.}Table~\ref{example:flashback} shows an example of a refined plan wherein the narrative structure is changed through the use of flashbacks to previous time events. Full story is provided in Table~\ref{example:full_flashback}.
\begin{table*}[t]
    \small
    \begin{tabular}{m{0.95\linewidth}}
    \toprule
Premise: Our protagonist is a high school junior struggling with his mental health. He can’t keep up with his school work due to his constant state of anxiety and feels he is failing at everything. He doesn’t know anyone with similar struggles and feels lonely. One day, he finds an envelope with \$500 cash in it when he is walking with his dog in the forest behind his house. This event becomes a turning point when he meets “Abe,” a homeless man, living in the forest.\\
\\
Settings: The story is set in a small town in the American Midwest.\\
\\
Characters:\\
Alyssa Brown: Alyssa Brown is 17-year-old high school classmate of John Smith's, who struggles with mental health issues herself but manages to reach out to him and become his friend.\\
Abe: Abe is 45-year-old homeless man who has been living in the forest for over 10 years, and who John befriends after finding the mysterious package of cash.\\
John Smith: John Smith is 17-year-old high school student who stumbles upon something mysterious while walking his dog in the woods.\\
Jake Jackson: Jake Jackson is 17-year-old new student at John Smith's school, and a talented artist who becomes friends with both John and Alyssa.\\
\\
Outline:\\
1. John discovers the package of cash and visits Abe every day to share his lunch, and they develop a bond. Scene:  Characters: Abe, John Smith\\
\\
\quad    a. John discovers the package of cash one day while walking his dog Max in the forest behind his house. Scene: the town of Millfield, Ohio, where John has lived his whole life. Characters: John Smith\\
\\
\quad    b. John visits Abe every day to share his lunch, and they develop a bond. Scene: the forest behind John's house. Characters: Abe, John Smith\\
\\
\quad    c. Abe decides to return the favor and repay John for his kindness. Scene: the forest behind John's house. Characters: Abe, John Smith\\
\\
2. With Abe's guidance, John starts an art therapy group at school, where he can connect with his peers and express himself creatively. Scene:  Characters: Abe, John Smith\\
\\
\quad    a. Abe guides John to start an art therapy group at school as a way to connect with his peers and express himself creatively. Scene: the school. Characters: Abe, John Smith\\
\\
\quad    b. John struggles to convince his classmates and school staff to allow the art therapy group to form, but eventually gains their support after sharing his own struggles with mental health. Scene: the school. Characters: John Smith\\
\\
\quad    c. John is able to find success and fulfillment as he becomes less anxious and more confident in himself. Scene: the school. Characters: John Smith\\
\\
3. While John's mental health improves, he becomes closer to Alyssa and Jake, who support him in exploring his artistic side and his aspirations. Scene:  Characters: Alyssa Brown, John Smith, Jake Jackson\\
\\
\quad    a. John forms a growing friendship with Alyssa through their connection over mental health issues. Scene: the school. Characters: Alyssa Brown, John Smith\\
\\
    b. John starts to spend more time with and draw closer to Alyssa as he navigates his renewed sense of self and purpose. Scene: the school. Characters: Alyssa Brown, John Smith\\
\\
\quad    c. The art therapy group brings John closer not only to his classmates but also to Jake, another talented artist in the class. Scene: the school. Characters: Jake Jackson, John Smith\\

    \bottomrule
    \end{tabular}
    \caption{Example of a refined plan wherein the narrative structure is enhanced through the use of flashbacks to previous time events.}
    \label{example:flashback}
\end{table*}
\begin{table*}[t]
    \small
    \begin{tabular}{m{0.95\linewidth}}
    \toprule
On a sunny afternoon, as the school day came to a close, John found himself sitting in the art therapy group that he had worked so hard to create. The atmosphere was filled with an air of vulnerability as the members opened up, sharing their deepest fears and struggles through various creative mediums.

John had always been hesitant to reveal his own battles with anxiety and depression, but today felt different. Today, he knew it was time to let go of his insecurities and share his story.

Taking a deep breath, John stood up in front of the group. His hands trembled slightly as he held up a canvas painted with swirls of dark blues and stormy grays . The painting reflected the chaos and turmoil within his own mind, a visual representation of the constant battle he faced.

"I've always felt like I was drowning in my own thoughts," John began, his voice wavering slightly. "Anxiety has been a suffocating presence in my life, making it impossible to focus on anything else. It's like a storm that rages inside me, constantly churning and swirling, making it hard to find any sense of calm."

As John spoke, his gaze shifted from the canvas to the faces of the members in his art therapy group. Alyssa sat nearby, her eyes filled with understanding and compassion. Jake, his childhood friend, leaned forward attentively, offering a reassuring smile that conveyed both support and admiration.

The room fell into a hushed silence as John continued to reveal the inner turmoil that had plagued him for so long. The vulnerability he displayed in sharing his struggles resonated deeply with the group members , many of whom had faced their own battles with mental health. Each brushstroke in John's painting seemed to capture the weight of his anxiety and depression, the tangled web of thoughts and worries that consumed him.

As he finished speaking, a wave of emotions washed over John. He felt a mix of relief and trepid ation, unsure of how the group would react to his vulnerability. But as he looked around the room, he saw understanding in their eyes. They had all experienced their own struggles with mental health, and John's words resonated deeply within them.

Alyssa was the first to break the silence. Her voice trembled as she spoke, her eyes filled with empathy. "John, thank you for sharing. Your words are powerful, and your vulnerability is a strength."

Jake nodded in agreement, his normally lighthearted expression replaced with sincere concern. "You're not alone in this battle, man," he said softly. " We're all here for you, every step of the way."

John felt a lump forming in his throat as he absorbed their support. He never expected such understanding and compassion from his classmates. This art therapy group, which he initially joined out of desperation, had become a lifeline for him—a place where he could express his deepest struggles and find solace in the shared experiences of others.

As the weeks went by, John's mental health steadily improved. The therapeutic process of creating art allowed him to channel his anxiety and insecurities into something beautiful and meaningful. With each stroke of paint on canvas or pencil sketch on paper, he felt a sense of release and a renewed sense of purpose. The art therapy group became a space for him to not only express himself but also find support and understanding from his peers.

Alyssa, with her own battles against mental health, became John's rock throughout this journey. She understood the weight he carried on his shoulders and was always there to listen and offer words of encouragement. Their friendship blossomed as they spent more time together, bonding over their shared experiences and dreams for the future.

One day, as John and Alyssa sat on a park bench, discussing their latest art projects, Jake approached them with a mischievous smile. He held up a flyer for a local art exhibition. "Hey guys," he said, his excitement evident in his voice, "I found this and thought it might interest you. It's an open call for artists to showcase their work at the town gallery next month."

John's heart skipped a beat as he scanned the flyer and saw the colorful images of various artworks that adorned its surface. The vibrant brushstrokes and intricate details ignited a spark within him, reminding him of his passion for art that he had almost forgotten amidst his struggles with mental health.

Alyssa's eyes lit up with equal enthusiasm as she leaned in closer to examine the paintings on the flyer. Her fingers traced along the edges of a particularly striking piece, her excitement growing alongside John's.

"This is amazing," Alyssa said, her voice barely able to contain her joy. "We should definitely submit our work. It could be a chance to showcase our art and connect with other artists who understand us."

John nodded, a smile tugging at the corners of his lips. The thought of displaying his art in a public exhibition filled him with both anticipation and anxiety. But he trusted Alyssa's words and believed in the power that art held to transcend language and communicate emotions.

As they walked down the school hallway, John couldn't help but reflect on how far he had come. It felt like a lifetime ago when he discovered that envelope filled with money in the forest behind his house. The memory started to play in his mind like an old movie reel.

It was a cloudy day, the air heavy with anticipation of an approaching storm. John, lost in his thoughts and struggling with the weight of his anxiety, took Max for a walk in their usual spot – the forest behind his house. As they wandered deeper into the woods, Max's tail wagging happily, John's eyes caught sight of something shiny among the leaves.

Curiosity overcame his anxiety, and he knelt down to get a closer look. He reached out and brushed aside the fallen leaves, revealing an envelope. His heart raced with anticipation as he opened it to find \$500 in cash neatly stacked inside.

At that moment, John's mind became a whirlwind of questions. Who had left this money here? Why? And what should he do with it? He had never seen such a large amount of money before, and the thought of keeping it made his heart race even faster. But something inside him told him that this was an opportunity, a chance to make a difference in his own life and the lives of others.

John's hands trembled as he stared at the stack of money. The weight of his anxiety lifted slightly, overshadowed by the possibilities that lay before him. With the cash in his possession, he could alleviate some of the financial burden on his struggling family. He could finally invest in art supplies that he had been longing for, expanding his creative abilities. ...
\\
    \bottomrule
    \end{tabular}
    \caption{Example of a narrative employing a flashback structure to weave an extended story.}
    \label{example:full_flashback}
\end{table*}
\subsection{Anthropomorphism.} Table~\ref{exam:personalize} shows an example of refined plan wherein the protagonist’s loneliness is anthropomorphized, enhancing the uniqueness of narrative setting. Full story is provided in Table~\ref{example:full_persona}.
\begin{table*}[t]
    \small
    \begin{tabular}{m{0.95\linewidth}}
    \toprule
    Premise: After a long day at work, the last thing Aimee wants is to go home to her empty apartment. She's tired of being alone, but she can't seem to find anyone who understands her. When she's about to give up, she meets a man who finally makes her feel understood.\\
\\
Setting: The story is set in Aimee's apartment.\\
\\
Characters:\\
Brianne Patterson: Brianne Patterson is a woman in her early twenties.\\
Jessica Sanders: Jessica Sanders is a woman in her late twenties.\\
Aimee Kincaid: Aimee Kincaid is a young woman in her early twenties.\\
David Kwan: David Kwan is a young boy in his early teens.\\
Kyle Johnson: Kyle Johnson is a man in his early thirties.\\
Patricia Hill: Patricia Hill is an elderly woman in her seventies.\\
Loneliness (personified): Loneliness is a physical entity that Aimee can interact with, representing her feelings of loneliness and isolation.\\
\\
Outline:\\
\\
1. Aimee Kincaid goes home after a long day at work only to find her apartment empty and her loneliness manifests as a physical entity. Scene:  Characters: Aimee Kincaid, Loneliness(personified)\\
    \\
\quad    a. Aimee Kincaid enters her empty apartment after an exhausting day at work and encounters Loneliness, a physical manifestation of her solitude. Scene:  Characters: Aimee Kincaid, Loneliness(personified)\\\\
\quad    b. Loneliness interacts with Aimee, amplifying her feelings of isolation and despair. Scene:  Characters: Aimee Kincaid, Loneliness(personified)
\\\\
2. Kyle Johnson, a colleague from work, visits Aimee, interrupting her interaction with Loneliness and befriending her. Scene:  Characters: Kyle Johnson, Aimee Kincaid, Loneliness(personified)\\
\\
\quad    a. Kyle Johnson comes over to Aimee's apartment for a visit, breaking the solitude that Aimee and Loneliness had been in. Scene:  Characters: Kyle Johnson, Aimee Kincaid, Loneliness(personified)\\\\
\quad    b. Aimee and Kyle swiftly establish a friendship, sparked by their shared affection for books and movies. Scene:  Characters: Aimee Kincaid, Kyle Johnson\\
\\
3. Aimee trusts Kyle enough to share her experience of interacting with her physical loneliness, expressing her fear of solitude. Scene:  Characters: Kyle Johnson, Aimee Kincaid, Loneliness(personified)\\
    \\
\quad    a. Aimee courageously shares her fears and experiences of Loneliness with Kyle, divulging her dread of remaining alone. Scene:  Characters: Kyle Johnson, Aimee Kincaid\\\\
\quad    b. With a sympathetic ear, Kyle confesses that he sometimes feels the same way, deepening their connection. Scene:  Characters: Kyle Johnson, Aimee Kincaid\\
\\
4. Reassured by Kyle's understanding and empathy, Aimee experiences a moment of solace and companionship, changing her perspective on her manifested loneliness. Scene:  Characters: Kyle Johnson, Aimee Kincaid, Loneliness(personified)\\
    \\
\quad    a. Kyle consoles Aimee, leading to an intimate conversation that gives Aimee a sense of understanding and companionship. Scene:  Characters: Kyle Johnson, Aimee Kincaid\\\\
\quad    b. After their heart-to-heart, Aimee realizes that with companionship, Loneliness loses its strength, giving her a new perspective as they part ways. Scene:  Characters: Aimee Kincaid, Loneliness(personified)\\
    \bottomrule
    \end{tabular}
    \caption{Example story plan of  enhancing originality of story setting by personifying the loneliness of the story's characters through \RePlan{}.}
    \label{exam:personalize}
\end{table*}
\begin{table*}[t]
    \small
    \begin{tabular}{m{0.95\linewidth}}
    \toprule
     Aimee trudged through the bustling streets of the city, her tired feet dragging along the pavement. Her mind was weary from a long day at work, each minute feeling like an hour as she navigated through endless meetings and demanding deadlines. As the sun started to set, casting its golden glow across the city , Aimee couldn't help but notice the beauty of the sky. The vibrant hues of orange and pink painted a breathtaking scene above her.

Finally reaching her apartment building, Aimee climbed up the worn-out stairs, her anticipation growing with each step. But as she opened the door to her small apartment, she was met with an overwhelming emptiness that seemed to fill the air. The silence enveloped her like a heavy cloak, amplifying the ache in her heart. Aimee had grown tired of returning home to the same empty space day after day, feeling as though she was missing something vital in her life.

As she stood in the midst of her silent apartment, Aimee couldn't shake off the feeling that there was an eerie presence accompanying her. It wasn't a physical presence per se, but rather an intangible entity that seemed to lurk in the corners of her vision. Loneliness.

Loneliness, personified as a shadowy figure with long, wispy tendrils that reached out towards Aimee, exuded an aura of melancholy. Its form was constantly shifting, appearing as a haunting silhouette one moment and dissipating into thin air the next. Yet, there was no mistaking its presence.

Aimee felt her heart race as she stepped further into her apartment, the front door clicking shut behind her. The dim lighting created an atmosphere of uncertainty, casting long shadows across the room. She hesitated for a moment, her eyes fixated on the ethereal figure before her.

Loneliness.

The name echoed in Aimee's mind as she stood frozen, her body trembling with a mix of fear and curiosity. Loneliness seemed to study her, its ghostly presence holding an unspoken invitation. Aimee took a tentative step forward, drawn to the figure despite her apprehension.

The room had transformed in the presence of Loneliness. The once cozy and familiar living space now felt cold and desolate. The air hung heavy with an oppressive silence, the vibrant colors of her belongings muted into shades of gray. Aimee could feel the weight of her loneliness pressing down on her chest, suffocating her.

Loneliness moved closer, its wispy tendr ils swaying in the stillness of the room. Aimee's heart pounded in her chest, her breaths coming in shallow gasps. She could see Loneliness's form shifting and swirling like smoke, as if it were made of mist and shadow.

As Aimee stood frozen in place, Loneliness extended a ethereal hand towards her, its touch sending shivers down her spine. Aimee couldn't tear her eyes away from the figure before her - the embodiment of all her pain and longing.

In a voice that seemed to resonate within Aimee's very soul, Loneliness whispered words that pierced through the silence of the apartment. "You will always be alone. No one will ever truly understand you," it hissed, its voice echoing with a haunting emptiness.

Aimee felt her heart ache as Loneliness's words seemed to confirm her deepest fears. She had spent countless nights searching for someone who could see beyond her facade and truly understand her, but each time she found herself more disappointed and disillusioned. Loneliness had become her constant companion, a cruel reminder of her isolation.

But just as Aimee began to lose hope, there was a soft knock on the door. Startled, she turned toward the entrance of her apartment, her eyes still fixated on Loneliness. She hesitated for a moment, her heartbeat pounding in her ears. Who could it be? And why now, when she was at the lowest point of her despair?

Taking a deep breath, Aimee slowly made her way to the door. She gently placed her hand on the doorknob, feeling a mix of apprehension and hope swirling inside her. With a shaky hand, she turned the knob and pulled open the door.

There stood Kyle Johnson, his genuine smile warming Aimee's heart. His wavy chestnut hair was slightly tousled from the wind, and his bright blue eyes sparkled with kindness. Aimee couldn't help but let out a small gasp of surprise and relief. It was as if the universe had finally heard her plea for companionship and sent Kyle right to her doorstep.

"Hey, Aimee," Kyle greeted her, his voice filled with warmth. "I hope I'm not interrupting anything."

Aimee shook her head, a genuine smile tugging at the corners of her lips. "No, not at all," she replied, her voice laced with newfound hope. "Actually, you couldn't have come at a better time."

Kyle's smile widened, mirroring the growing excitement in Aimee's eyes. "I'm glad to hear that," he said sincerely. "I was thinking about grabbing dinner nearby, and I remembered how you mentioned this new Italian restaurant. Would you like to join me?"

Aimee's heart flutter ed at the invitation. She couldn't believe her luck. It felt like a dream come true, having someone to share a meal and lively conversation with after such a long time of solitude.

"Yes! I'd love to," she replied, feeling a surge of happiness that radiated through her entire being.

As they stepped out of the apartment, Aimee couldn't help but notice the change in the atmosphere. The evening air felt lighter, filled with a sense of possibility and connection that she hadn't felt in a long time. She glanced back at her apartment, where Loneliness stood silently at the threshold. Aimee smiled softly as she locked the door behind her, feeling a weight lifted off her shoulders. She knew that Loneliness would always be there, just like an old familiar friend, but tonight was different. Tonight, she had someone who understood.

As Aimee walked alongside Kyle down the dimly lit streets of the city, her heart felt light and her steps were a little quicker than usual. The soft glow of the street lamps illuminated their path, casting a warm and comforting ambiance around them.

Aimee glanced at Kyle, who seemed genuinely interested in everything she had to say. It was as if he had a genuine curiosity about her thoughts and perspectives, and Aimee found herself opening up to him in a way she hadn't with anyone else before. The conversation flowed effortlessly between them, a seamless exchange of ideas and stories.
...\\
    \bottomrule
    \end{tabular}
    \caption{Example of a narrative that uses an anthropomorphized character to tell an extended story.}
    \label{example:full_persona}
\end{table*}
\subsection{Complex Character Relation.}
Table~\ref{example:complex_relation} shows an example of complex network of character relationships within an extended narrative.
\begin{table*}[t]
    \small
    \begin{tabular}{m{0.95\linewidth}}
    \toprule
 Jake sat on his bed, rubbing the sleep from his eyes as he stared at the unfamiliar message illuminating his phone screen. The words were ominous, their weight sinking into him with a sense of urgency. "If you want to live – listen!"

His pulse quickened, and he read the message again, hoping for some clue as to who could be behind it. But there was nothing. The sender was unknown, and the message itself offered no explanation.

As Jake's mind raced with possibilities, he couldn't shake the feeling that he had stumbled upon something far bigger than himself. It was like stepping into a world of mystery and danger that he couldn't comprehend. Determined to uncover the truth, Jake reached out to his closest friends, Sarah and Tom, knowing that their unique skills and unwavering loyalty would be crucial in navigating this treacherous journey.

Together, they formed an unlikely trio of adventure-seekers, each bringing their own expertise to the table. Jake, a master of puzzles and riddles, had always been fascinated by hidden treasures and secret codes. Sarah, a skilled hacker with an uncanny ability to dig up information, was invaluable when it came to uncovering clues and untangling webs of deceit. And Tom, a fearless adrenaline junkie who thrived on danger, was always the first to charge into unknown territory. Together, they made an unbeatable team.

As the trio gathered around the table in their makeshift headquarters, a rundown attic filled with old books and dusty artifacts, Jake spread out a worn map that he had discovered during their investigation. It depicted a labyrinthine series of tunnels beneath the city, leading to what they believed to be the location of the long-lost treasure. The map was intricate, with dotted lines and cryptic symbols that Jake had spent countless hours deciphering.

"We're getting closer," Jake said, excitement bubbling in his voice. "This is it. The final piece to the puzzle. The treasure is within our reach."

Sarah leaned in closer, her eyes fixed on the map. "But we're not the only ones searching for it," she warned. "We have to be careful."

Tom scoffed, his confidence unwavering. "They wouldn 't stand a chance against us. We're the best team there is."

Emma chimed in, her voice filled with excitement. "And with this map, we have the upper hand. It's like a labyrinth of secrets waiting to be unraveled."

Jake nodded, his eyes gleaming with determination. "That's right. We've come too far to turn back now. We're going to find that treasure and no one is going to stop us."

Sarah's gaze met Jake's, a mixture of worry and affection in her eyes. "Just promise me you'll be careful, Jake. This treasure seems to have brought out the worst in people. I don't want anything bad to happen to you."

Jake reached out and took Sarah's hand, giving it a reassuring squeeze. "I promise, Sarah. We'll all watch each other's backs and make sure we come out of this adventure unscathed."

Tom smirked, crossing his arms over his chest. "Well, well, it's nice to see you two lovebirds sharing a moment," he chuckled.

Sarah shot Tom an annoyed look before turning her attention back to Jake. "Just be careful, okay? I can't bear the thought of anything happening to you."

Jake smiled at Sarah, grateful for her concern. "Don't worry, Sarah. I promise I'll be careful," he reassured her. "We've come too far to back down now."

Tom rolled his eyes and let out a mock sigh. "Oh please, Sarah, it's not like we're going up against an army here. We can handle whatever comes our way."

Sarah shot Tom a glare before turning her gaze back to Jake. "Fine, but still, just be cautious. We don't know what kind of dangers lie ahead."

Jake nodded in agreement. "You're right, Sarah. We need to stay vigilant and think carefully about our next moves. We can't afford to make any mistakes now."

With their plan in motion, the trio ventured deeper into the belly of the city. The air was thick with anticipation as they approached their final destination. The stakes were higher than ever before, and their adrenaline surged through their veins.

As they rounded a dark corner, they came face to face with a steel door. Its ominous presence sent shivers down their spines, but they knew that beyond it lay the ultimate prize they had been tirelessly pursuing.

Jake took a deep breath and reached for the handle, his heart pounding in his chest. With a firm grip, he twisted the door handle and pushed it open, revealing a hidden chamber bathed in golden light. Their eyes widened in awe as they took in the sight before them.

The chamber was filled with ancient artifacts, sparkling jewels, and stacks of valuable artwork. It was a treasure trove beyond their wildest imaginations. As Jake, Sarah, and Tom stepped into the hidden chamber, their jaws dropped in amazement. The golden light dancing off the precious artifacts created an ethereal glow that enveloped them. They could hardly believe their luck – they had stumbled upon a treasure trove of immense value.

Ancient Egyptian statues stood tall amongst ornate jewels and intricately designed pottery. The walls were adorned with mesmerizing paintings, depicting scenes of ancient civilizations and mythical creatures. Each step they took seemed to reveal another awe-inspiring artifact, further confirming that they were standing in the presence of history's greatest treasures.

But as their amazement subs ided, a sense of urgency started to take hold. They knew they couldn't waste any time basking in the splendor of the chamber. The rival factions were hot on their heels, and danger lurked around every corner ...\\
    \bottomrule
    \end{tabular}
    \caption{Example of a complex network of character relationships within an extended narrative.}
    \label{example:complex_relation}
\end{table*}
\subsection{Human-Machine Interactive Writing.}\label{ex:inter} 
\begin{figure*}[t]
    \centering \includegraphics[width=2.0\columnwidth]{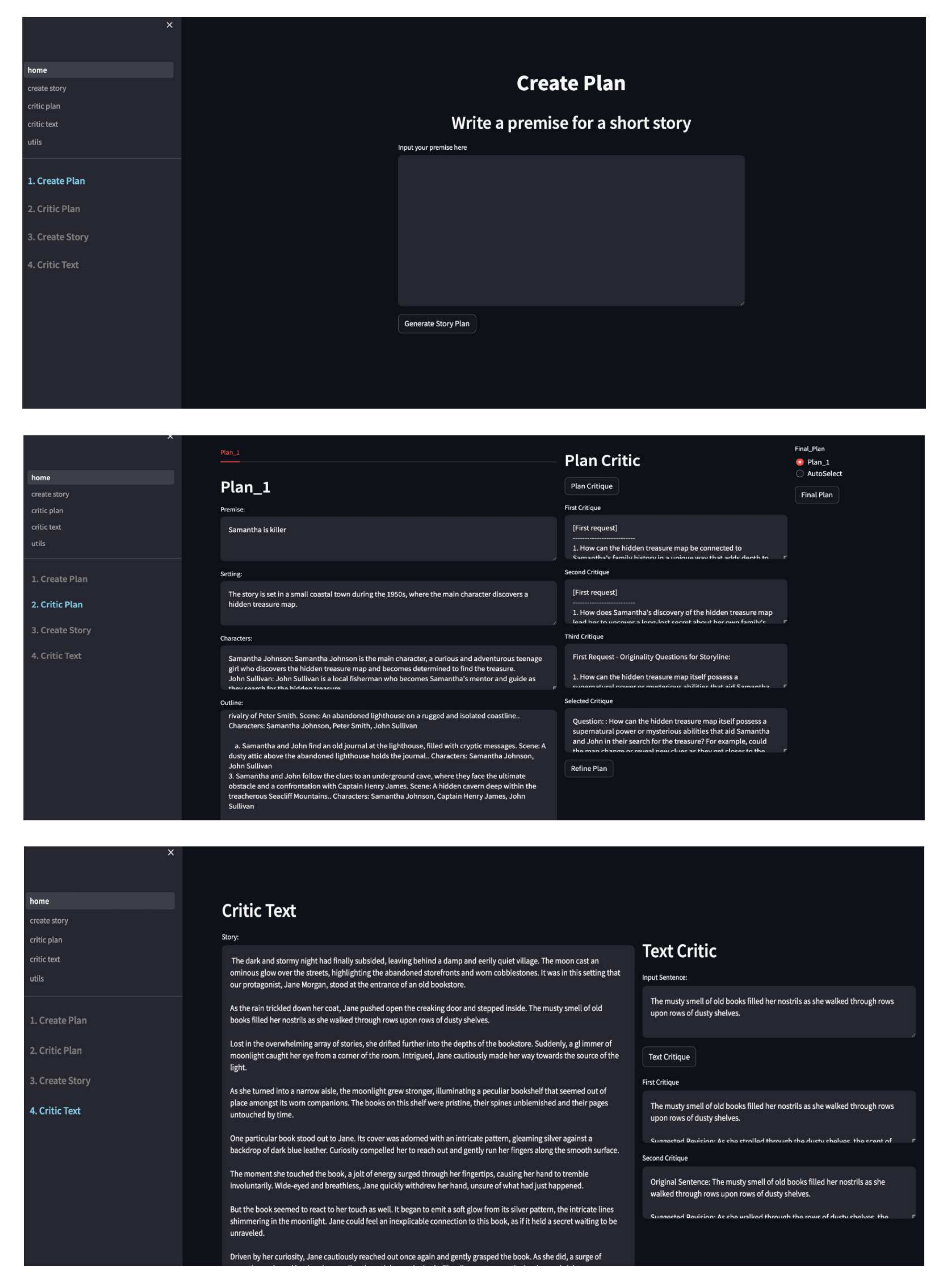}
    \caption{Web application for human-machine interactive writing.}
    \label{fig:interface-inter}
\end{figure*}

Table~\ref{exp:inter_plan} and Table~\ref{exp:inter_story} show examples of narratives generated through human-machine interactive writing, created based on the premise of `a baby skeleton riding a skateboard.'
During the critique process in the \RePlan{} stage, a human user add new critique content related to skateboarding tricks, which leads to the inclusion of skateboarding elements in the story. In the \ReSent{} stage, the plain expression ``vintage skateboards and photographs, capturing his daring stunts'' is revised to the more expressive ``chock-full with retro boards, killer shots of his gnarly stunts.'' These changes enhance the tone and style of the narrative context, making it more vivid and dynamic. Additionally, we develop a web interface to facilitate this interactive writing process, demonstrating its usability and effectiveness in real-world scenarios, as illustrated in Figure~\ref{fig:interface-inter}.
\begin{table*}[t]
    \small
    \begin{tabular}{m{0.95\linewidth}}
    \toprule
    Premise: A baby skeleton riding a skateboard\\
\\
Setting: The story is set in a small, forgotten skate park tucked away in the heart of an abandoned amusement park.\\
\\
Characters:
Baby Bones: Baby Bones is the baby skeleton who comes to life and rides a skateboard in the skate park.\\
Sammy Skater: Sammy Skater is a young, adventurous teenager who discovers Baby Bones at the abandoned amusement park.\\
Shimmer: Shimmer is a mischievous ghost who haunts the skate park and befriends Baby Bones.\\
Maxine "Max" Decker: Maxine "Max" Decker is a former professional skateboarder who now runs a skateboarding shop near the amusement park.\\
\\
Outline:\\
1. Sammy Skater discovers the abandoned amusement park and encounters Baby Bones, a skateboard-riding skeleton who springs to life. Scene: A decrepit, overgrown skate park shrouded behind rusty gates. Characters: Baby Bones, Sammy Skater\\
\\
\quad    a. Sammy Skater stumbles upon the rusty gates of the abandoned amusement park, intrigued, he decides to explore. Scene: A secluded and overgrown forest filled with haunting murmurings. Characters: Sammy Skater, Baby Bones\\
\\
2. Shimmer, a visible spectral entity, befriends Baby Bones and unravels the park's haunted past, explaining the tragic skateboard accident of Maxine Decker's brother, the celebrated skateboarder. Scene: The dimly lit, eerily silent basement hallways of the amusement park. Characters: Baby Bones, Shimmer, Sammy Skater\\
\\
\quad   a. Shimmer presents Baby Bones with a timeworn newspaper article regarding the ominous accident at the skate park. Scene: The dusty attic in the dilapidated amusement park building, filled with forgotten memorabilia. Characters: Baby Bones, Shimmer, Sammy Skater, Ghost of Max Decker\\
   \\
\quad   b. Shimmer and the Ghost of Max Decker reveal to Baby Bones the phenomenal legacy of Maxine Decker's late brother. Scene: An ancient memorial brimming with skateboard being slowly swallowed by dust and spiderwebs. Characters: Shimmer, Ghost of Max Decker, Baby Bones\\
\\
3. The ghost of Max Decker takes on the role of a mentor, helping Baby Bones to uncover remarkable, gravity-defying skateboarding tricks that gives the illusion of flying. Scene: The ghostly and ethereal silhouette of the abandoned skate park under an eerie moonlit sky. Characters: Ghost of Max Decker, Baby Bones\\
\\
\quad   a. Guided by the ghost of Max Decker, Baby Bones begins to learn and perfect death-defying tricks on the skateboard, gradually revealing an innate talent. Scene: The ominous but captivatingly beautiful abandoned amusement park under the spectral gleam of the full moon. Characters: Baby Bones, Ghost of Max Decker, Sammy Skater, Shimmer\\
\\
4. Inspired and supported by Sammy, Shimmer, and the guiding spirit of Max Decker, Baby Bones challenges the status quo at a skateboarding competition, baffling everyone present with their unearthly talent. Scene: A lively city skate park bustling with graffiti artists and skateboarders. Characters: Baby Bones, Sammy Skater, Shimmer\\
\\
\quad   a. Baby Bones, performing eerie tricks that seem almost spectral, wins the stunned silence of the crowd. Scene: A city-wide skateboarding competition swarming with onlookers spying on every move with hawk-like vigilance. Characters: Baby Bones, Sammy Skater, Shimmer\\
\quad   b. The crowd bursts into awe-struck applause as Baby Bones flawlessly delivers a series of mystical, gravity-defiant tricks. Scene: A massive stadium housing a renowned international skateboarding championship. Characters: Baby Bones, Sammy Skater, Shimmer, Ghost of Max Decker\\
\quad   c. Baby Bones honored as the champion of the skateboarding competition, their spectral talent bringing both fear and jubilation to the spellbound spectators. Scene: A colossal outdoor skate park vibrating with the roar of an awestruck crowd. Characters: Baby Bones, Sammy Skater, Shimmer, Maxine "Max" Decker.\\
    \bottomrule
    \end{tabular}
    \caption{Example of story plan generated by human-machine interactive writing.}
    \label{exp:inter_plan}
\end{table*}
\begin{table*}[t]
    \small
    \begin{tabular}{m{0.95\linewidth}}
    \toprule
     The dark and stormy night had finally subsided, leaving behind a damp and eerily quiet village. The moon cast an ominous glow over the streets, highlighting the abandoned storefronts and worn cobblestones. It was in this setting that our protagonist, Jane Morgan, stood at the entrance of an old bookstore.
As the rain trickled down her coat, Jane pushed open the creaking door and stepped inside. The musty smell of old books filled her nostrils as she walked through rows upon rows of dusty shelves.
Lost in the overwhelming array of stories, she drifted further into the depths of the bookstore. Suddenly, a gl immer of moonlight caught her eye from a corner of the room. Intrigued, Jane cautiously made her way towards the source of the light.
As she turned into a narrow aisle, the moonlight grew stronger, illuminating a peculiar bookshelf that seemed out of place amongst its worn companions. The books on this shelf were pristine, their spines unblemished and their pages untouched by time.
One particular book stood out to Jane. Its cover was adorned with an intricate pattern, gleaming silver against a backdrop of dark blue leather. Curiosity compelled her to reach out and gently run her fingers along the smooth surface.
The moment she touched the book, a jolt of energy surged through her fingertips, causing her hand to tremble involuntarily. Wide-eyed and breathless, Jane quickly withdrew her hand, unsure of what had just happened.
But the book seemed to react to her touch as well. It began to emit a soft glow from its silver pattern, the intricate lines shimmering in the moonlight. Jane could feel an inexplicable connection to this book, as if it held a secret waiting to be unraveled.
Driven by her curiosity, Jane cautiously reached out once again and gently grasped the book. As she did, a surge of warmth envelop ed her hand, spreading through her entire body. The silver pattern on the book grew brighter, illuminating the surrounding area with an otherworldly glow.
With newfound confidence, Jane opened the book to its first page. Words appeared before her eyes, written in an elegant script that seemed to dance across the page:
"The untold history of the skate park awaits those brave enough to uncover its secrets."
\\
Intrigued and feeling a sense of destiny, Jane turned another page and was greeted by a vivid illustration depicting the skate park in its former glory, full of laughter and excitement. She couldn't help but notice the figure of a young girl, her face glowing with joy as she skated through the park. Jane's heart filled with an inexplicable longing, as if she somehow knew this girl.\\
\\
As the image on the page seemed to come alive, Shimmer's voice echoed in Jane's mind. "That was me," she whispered, her voice filled with a mix of nostalgia and sorrow. "I was once a skater here, just like you Sammy," she confessed, her eyes brimming with emotion.
Sammy Skater's eyes widened in surprise as the realization sank in. Shimmer was not just any ghost haunting the skate park; she was its heart and soul, forever connected to the memories and tragedies that had taken place within those old worn-out walls.
The dusty attic's floor creaked under their weight as Shimmer guided Baby Bones towards a weathered wooden crate. With trembling hands, she carefully opened it to reveal an assortment of faded newspaper clippings from decades past. The headlines screamed tales of triumph and tragedy, capturing the essence of a bygone era.
"Look," Shimmer whispered to Baby Bones, pointing at one particular article. The yellowed paper chronicled a tragic accident that had occurred at the skate park long ago. It detailed how a young girl named Maxine "Max" Decker had met her unfortunate demise while attempting a daring trick. Her loss had cast a dark shadow over the park, forever changing its vibrant atmosphere.
Baby Bones stared at the article, his empty eye sockets fixed on Max's photograph. He felt a deep sense of sorrow and empathy as if he had a connection to Max, even though they had never met. He could sense the weight of her tragedy and the impact it had on the skate park.
Shimmer watched Baby Bones closely, understanding that this revelation was hitting him hard. She gently placed a hand on his bony shoulder, offering comfort in this moment of unraveling the park's dark history. The dusty attic surrounded them, filled with nostalgia and forgotten memories.
"Baby Bones," Shimmer began softly, her voice laced with empathy, "the tragic accident at the skate park involved Maxine Decker's brother. He was a prominent skateboarder just like you , Baby Bones. His name was Jake Decker."\\
\\
Baby Bones's eyes widened beneath his tiny skeleton skull as Shimmer continued her somber revelation. "Jake was known for his fearless tricks and mesmerizing skills on the skateboard. He was adored by everyone in this skateboarding community, including Maxine."
The dusty skateboarding museum stood as a testament to Jake's legacy, filled with vintage skateboards and photographs capturing his daring stunts. Maxine Decker, Jake's older sister and owner of the vibrant skateboarding shop, stood beside Baby Bones, her eyes glistening with memories as she listened to Shimmer recount the tragedy of the skate park's past. Maxine could still vividly recall the day her brother disappeared, leaving behind a void in her heart that never truly healed.
As Shimmer spoke, Maxine's grip tightened on the skateboard in her hand. She had watched Jake grow into a fearless and skilled skateboarder who pushed boundaries and inspired everyone around him. She had admired his dedication and perseverance, often cheering him on from the sidelines as he conquered new tricks and soared through the air with grace.
But that fateful day at the skate park changed everything. Maxine's eyes brimmed with tears as she remembered the chaos that ensued when Jake attempted a daring trick on a ramp that went horribly wrong. She could still hear the sickening crack of Jake's bones as he collided with the concrete, and feel the weight of her heart as she rushed to his side, knowing deep down that this would be his last ride.
Maxine couldn't let go of her guilt. It had consumed her for years, ever since that fateful day when Jake disappeared at the skate park. As she watched Baby Bones and the others prepare for the skateboarding competition, memories of her younger brother flooded her mind.
She had always been protective of Jake, the fearless daredevil who never hesitated to push the boundaries of skateboarding. Maxine blamed herself for not being there that day, for not stopping him from attempting that fatal trick. If only she had been able to talk some sense into him, maybe he would still be here, alive and thriving.
As Baby Bones effortlessly glided across the ramps in Max ine's colorful backyard skate park, Maxine couldn't help but feel a mixture of awe and pain. The skeletal figure seemed to defy the laws of physics, soaring through the air with grace and precision. It was as if Baby Bones had tapped into a realm beyond mortal comprehension, where gravity held no power. ...
\\
    \bottomrule
    \end{tabular}
    \caption{Example of long story generated by human-machine interactive writing.}
    \label{exp:inter_story}
\end{table*}
\end{document}